\documentclass[final,5p,times]{elsarticle}

\usepackage{amsmath,amssymb}
\usepackage{bm}
\usepackage{booktabs}
\usepackage{array}
\usepackage{xcolor}
\usepackage{tikz}
\usetikzlibrary{arrows.meta,positioning,fit,backgrounds,calc}
\usepackage{subcaption}
\newcommand{\subfigref}[2]{\ref{#1}\subref{#2}}
\usepackage{placeins}   
\usepackage{stfloats}   
\usepackage{hyperref}   

\definecolor{cTrain}{HTML}{4878CF}   
\definecolor{cInput}{HTML}{E06C2B}   
\definecolor{cAlt}{HTML}{3A9E5C}     

\tikzset{
  block/.style={rectangle, rounded corners, draw=black!60, fill=cTrain!8,
    minimum height=9mm, minimum width=22mm, align=center, font=\small},
  frozen/.style={rectangle, rounded corners, draw=black!50, fill=black!12,
    minimum height=9mm, minimum width=22mm, align=center, font=\small},
  learn/.style={rectangle, rounded corners, draw=cTrain!75!black, fill=cTrain!18, very thick,
    minimum height=9mm, minimum width=22mm, align=center, font=\small},
  io/.style={rectangle, draw=black!45, fill=white, minimum height=8mm, align=center, font=\small},
  arr/.style={-{Stealth[length=2.2mm]}, thick},
}

\journal{Computer Vision and Image Understanding}

\begin{document}

\begin{frontmatter}
\title{3D Point Tracking with State Space Models}


\author[a]{Masahiro Ogawa\corref{cor1}}
\ead{ogawa@robot.t.u-tokyo.ac.jp}
\author[b]{Qi An}
\ead{anqi@robot.t.u-tokyo.ac.jp}
\author[b]{Atsushi Yamashita}
\ead{yamashita@robot.t.u-tokyo.ac.jp}
\cortext[cor1]{Corresponding author. Department of Precision Engineering, Graduate School of
Engineering, The University of Tokyo, 5-1-5 Kashiwanoha, Kashiwa, Chiba 277-8563, Japan.}
\affiliation[a]{organization={Department of Precision Engineering, Graduate School of Engineering,
  The University of Tokyo}, city={Kashiwa}, state={Chiba}, country={Japan}}
\affiliation[b]{organization={Department of Human and Engineered Environmental Studies,
  Graduate School of Frontier Sciences, The University of Tokyo},
  city={Kashiwa}, state={Chiba}, country={Japan}}

\begin{abstract}
Tracking any point of a dynamic scene in \emph{metric} 3D --- in absolute meters, not up to an
unknown scale --- underpins 3D and 4D reconstruction, robot navigation, and autonomous driving,
where decisions are made in meters, not pixels. Our objective is a 3D point tracker
accurate in those absolute terms and operating within a single commodity GPU, pose-free,
monocular budget. Our method rests on one observation: once a point's 2D image trajectory is fixed, the
quantity that governs its metric accuracy is the depth along its pixel ray. Rather than learning tracking end-to-end, we therefore compose
two frozen front-ends --- dense optical flow for 2D correspondence and a monocular metric-depth
network for the third dimension --- and learn only the residual they cannot supply: that depth, refined by a compact state space model (Mamba-3) conditioned on
appearance features (DINOv3). A state space model rather than the transformers the strongest 3D
trackers adopt is what makes a single-GPU budget attainable: it summarises a
track in a fixed-size recurrent state whose memory cost is constant in the number of frames,
whereas attention requires a key--value cache that grows linearly with them. On the TAPVid-3D minival
benchmark our best configuration attains the highest \emph{absolute} metric accuracy among
methods evaluated under identical conditions (mean metric Average Jaccard, $0.256$), exceeding
strong feed-forward trackers, while a companion analysis, reproduced with each competitor's own
evaluator, explains why several published trackers lose most of their accuracy under this budget.
\end{abstract}


\end{frontmatter}

\section{Introduction}
\label{sec:intro}

\subsection{Background}
\label{sec:intro-background}
Understanding a dynamic scene from images requires knowing not only \emph{where} a physical point on an
object appears in each frame's 2D image, but \emph{where it is in 3D} over time. This capability---tracking any point in
\emph{metric} 3D, in absolute meters rather than up to an unknown scale---underpins 3D and 4D reconstruction, robot navigation, and autonomous driving, where
decisions are made in meters rather than pixels. Recent 3D point trackers have made rapid progress on the
TAPVid-3D benchmark~\cite{tapvid3d}.

\subsection{Problem statement}
\label{sec:intro-problem}
Despite that progress, two obstacles limit the practical use of current 3D point trackers. First, the standard
leaderboard metric is \emph{scale-invariant}: it rescales each prediction by a single global factor before
scoring, which discards exactly the absolute-depth information that a downstream metric application needs.
A method can top the leaderboard while being metrically wrong. Second, the strongest published trackers
rely on large temporal windows~\cite{spatrackerv2}, precise camera poses~\cite{tapip3d}, or video-diffusion
backbones~\cite{trackcraft3r} that do not fit a commodity single-GPU budget; their reported numbers are
therefore not reproducible on the hardware most practitioners own.

\subsection{Objective}
\label{sec:intro-objective}
\begin{figure*}[!tb]
  \centering
  \begin{subfigure}[b]{0.50\linewidth}
    \includegraphics[width=\linewidth]{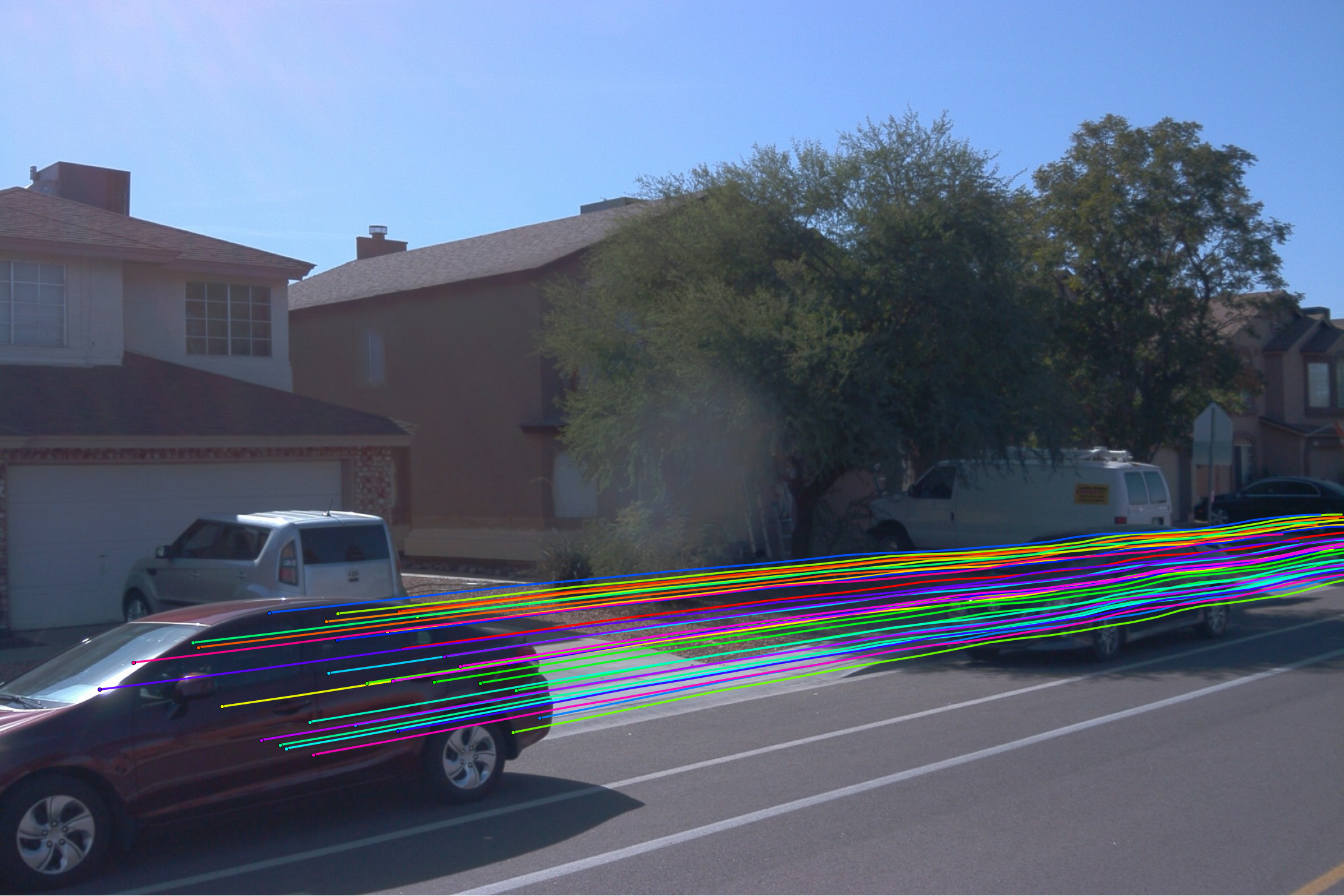}
    \caption{Tracked points on the input video}
  \end{subfigure}\hfill
  \begin{subfigure}[b]{0.359\linewidth}
    \includegraphics[width=\linewidth]{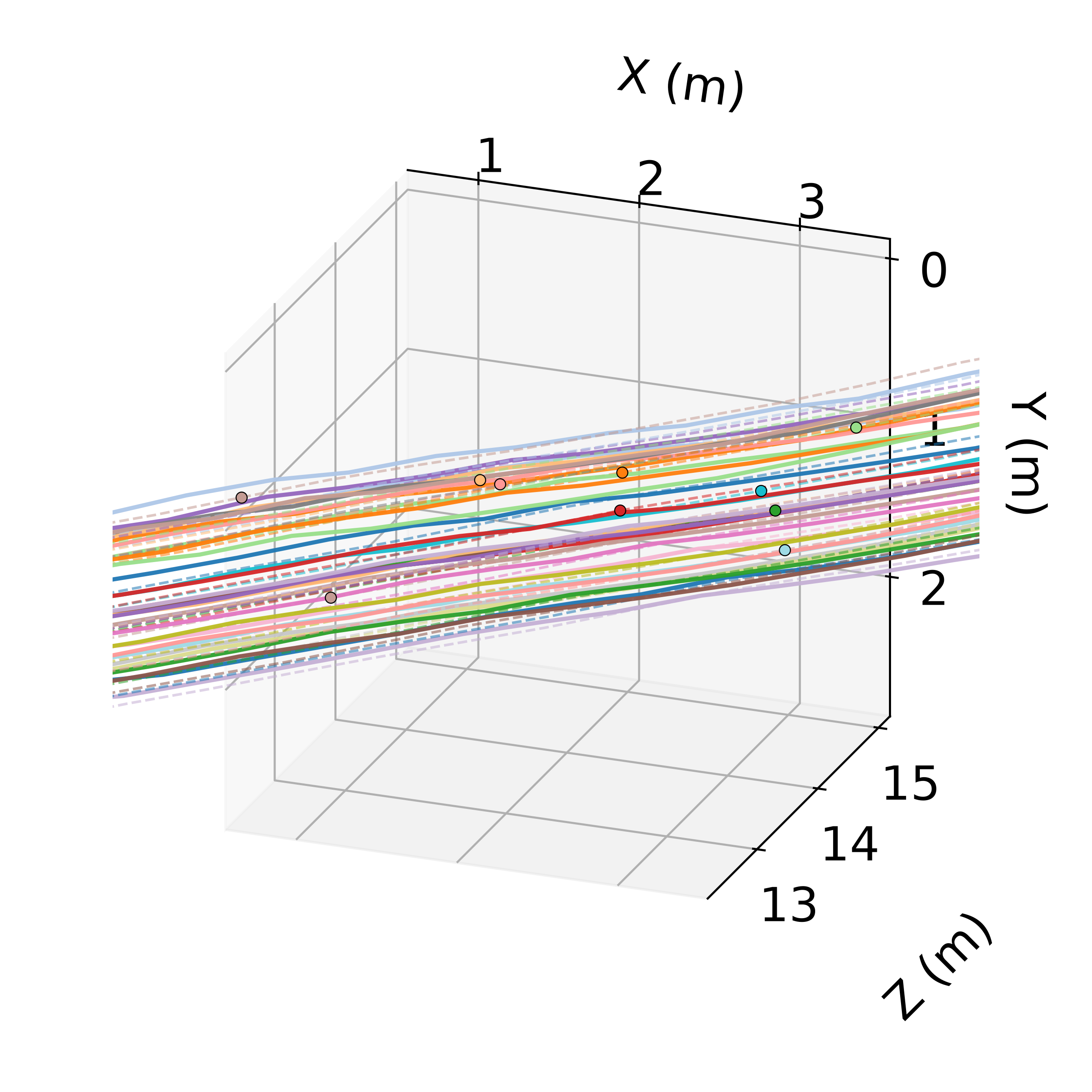}
    \caption{Recovered metric 3D tracks}
  \end{subfigure}
  \caption{Our metric 3D point tracking on a driving scene. (a) Query points are propagated in
  2D and overlaid on the video: each track is drawn as one curve from its start up to the frame
  shown, ending in a dot at the point's position in that frame; the panel is therefore the tracker's
  state at one instant rather than a summary of the whole clip. (b) The same points are lifted and
  refined into \emph{metric} 3D. The axis values are absolute meters in the DriveTrack~\cite{drivetrack}
  ground-truth frame, read off the data and not re-centred: $Z$ is distance from the camera, so this
  clip lies at $13$--$15$\,m, while $X$ and $Y$ are offsets about the optical axis and stay near zero.
  Solid = predicted, dashed = ground truth, dots =
  anchor frame; the view is cropped to a $3$\,m cube centred on the anchor points so that individual
  trajectories separate, and tracks that run outside it are clipped. Predicted tracks closely follow
  ground truth along
  all three axes $X$, $Y$, $Z$, not only in the $X$-$Y$ image-plane projection that a method with correct 2D
  tracking but wrong depth could also get right. Our depth-along-ray refiner corrects per-point depth so that the
  recovered tracks lie on the true surfaces in all three dimensions, not just when reprojected to 2D. The 3D
  view is oriented like the image of (a): $X$ to the right, $Y$ downward, $Z$ depth.}
  \label{fig:teaser}
\end{figure*}

Our objective is a 3D point tracker that is metrically accurate in \emph{absolute}, real-world terms and
reproducible on a single commodity GPU, without relying on the large temporal
windows, precise camera poses, or
heavy video-diffusion backbones that the strongest existing trackers need. Figure~\ref{fig:teaser} shows a
representative sample output of our system, tracking points in metric 3D on a real driving scene.

\subsection{Key idea}
\label{sec:intro-idea}
\begin{figure*}[!tb]
\centering
\resizebox{\textwidth}{!}{%
\begin{tikzpicture}[node distance=9mm and 12mm]
  \node[io] (vid) {video\\$\{I_t\}$};
  \node[frozen, right=of vid] (flow) {optical flow\\(frozen)};
  \node[frozen, right=of flow] (chain) {flow\\chaining};
  \node[io, right=of chain] (uv) {2D track\\$\mathbf{u}_t$ (+vis)};
  \node[frozen, below=17mm of flow] (da3) {metric depth\\(frozen)};
  \node[io, below=17mm of uv] (inp) {refiner inputs\\$[\,\mathbf{r}(\mathbf{u}_t),\ z_{\text{raw}},\ v\,]$};
  \node[learn, minimum width=30mm, minimum height=14mm, right=13mm of inp] (ref)
    {\textbf{learned}\\\textbf{depth refiner}\\{\footnotesize Figs.~\subfigref{fig:arch-ray}{fig:arch}, \subfigref{fig:v35-arch-concept}{fig:v35-arch}}};
  \node[io, right=13mm of ref] (out) {3D track\\$\hat{\mathbf X}_t$};

  \draw[arr] (vid) -- (flow);
  \draw[arr] (flow) -- (chain);
  \draw[arr] (chain) -- (uv);
  \draw[arr] (vid.south) |- (da3.west);
  \draw[arr] (uv) -- (inp);
  \draw[arr] (da3.east) -- (inp.west);
  \draw[arr] (inp) -- (ref);
  \draw[arr] (ref) -- (out);
\end{tikzpicture}%
}
\caption{Overall pipeline. The flow front-end (SEA-RAFT or WAFT) and the metric-depth backbone
(DA3-l or DA3-g, \S\ref{sec:eval}) are both frozen and swappable: swapping either only changes which
frozen network fills these two grey boxes, and re-training uses an identical architecture, only the
frozen depth cache changes. Flow chaining composes the front-end's consecutive-frame flow into a
full-length 2D track and, by a forward--backward consistency check, sets each point's visibility
$v$---the ``$+$vis'' of its output box (\S\ref{sec:track-2d}). Grey = frozen / training-free; the
\textbf{blue box} is the only learned module (under $1$M params); its internals are the Mamba-3
refiner (Fig.~\subfigref{fig:arch-ray}{fig:arch}) and the vmamba3 refiner (Fig.~\subfigref{fig:v35-arch-concept}{fig:v35-arch}). The Mamba-3
refiner leaves the frozen flow front-end's 2D track $\mathbf{u}_t$ (and visibility) untouched; the
vmamba3 refiner additionally applies a small bounded 2D correction to it.}
\label{fig:pipeline}
\end{figure*}
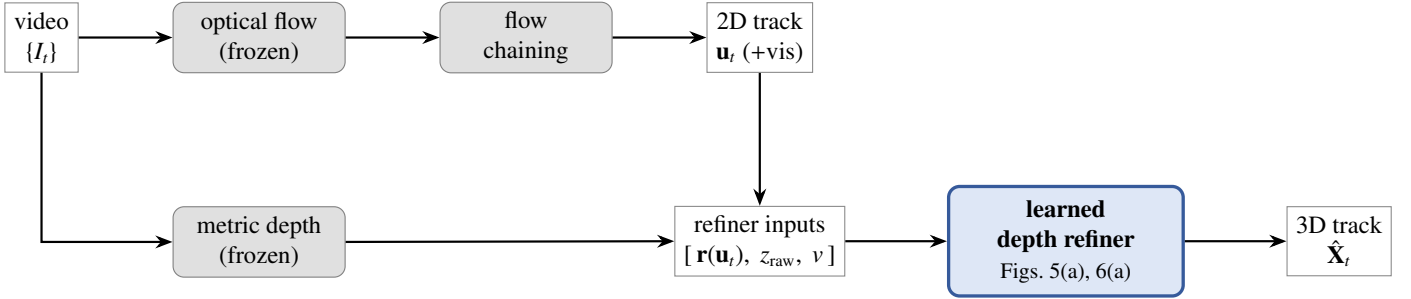

We take a deliberately modular and interpretable route. Our overall architecture is shown in
Fig.~\ref{fig:pipeline}. The one quantity that
governs metric 3D accuracy for a tracked point, once its 2D image trajectory is fixed, is the \emph{depth
along its pixel ray}. We
therefore compose two frozen, well-understood front-ends---dense optical flow~\cite{searaft,waft} for 2D
correspondence and a monocular metric-depth network~\cite{da3} (depth in absolute meters) for the third
dimension---and learn only
the depth residual they cannot supply, with a small refiner conditioned on
appearance features (DINOv3~\cite{dinov3}). The refiner runs a state space model
(Mamba-3~\cite{mamba3}) over the per-track temporal axis, mixing each track's frames at clip level;
it needs no spatial mixing of its own because its appearance and local-geometry inputs already carry
that spatial context---the DINOv3 feature is a whole-image summary at the tracked pixel, computed by
DINOv3's own frozen self-attention, and the local depth patch's samples occupy fixed grid positions
the network can learn to interpret directly. Applying such a model to image data requires removing
the causality that suits language but not images, which the method does as part of building the
refiner. We
abbreviate it \emph{vmamba3} in the prose and tables that follow, for readability, the leading \emph{v} standing
for \emph{vision}: it is Mamba-3 applied to image data.

We build the refiner on a state space model rather than a transformer for a concrete reason: it
summarises history in a \emph{fixed-size} recurrent state, whose memory cost is constant in the
sequence length, whereas softmax attention must retain a key--value cache that grows linearly with the
number of tokens. A 3D point track can span hundreds of frames of a long video, and this compact-state
property lets the refiner scale to such very long temporal sequences at a constant per-step memory
budget---a decisive advantage on the single commodity GPU we target, and one we quantify directly:
at a sequence length of $1025$ tokens the operator runs in $30$--$54\%$ of softmax attention's peak memory.

\subsection{Contribution}
\label{sec:intro-contrib}
Our contributions are:
\begin{enumerate}
  \item \textbf{A compact state space depth-along-ray 3D refiner} for metric 3D point tracking that
  corrects only per-point depth on top of frozen flow and monocular depth, keeping the learned module
  under $1$\,M parameters and the pipeline fully interpretable.
  \item \textbf{A depth scale refiner}: the larger of two metric-depth backbones is not
  simply better---it trades a learnable far-field bias for a frame-to-frame scale flicker no per-track
  refiner can observe---and we therefore stabilise the per-frame depth scale before refinement, reading
  the depth map directly.
  \item \textbf{VSSD-2pool, a second independently-read pool in the non-causal collapse.} Making Mamba-3
  non-causal costs its $\beta$-band: the band folds into a single per-token vector while the mask is
  causal, but not once that vector must also serve queries looking forward in the sequence, and
  recovering it there requires a second, independently-read pool. We derive, implement and evaluate that
  operator, and it raises absolute metric accuracy on TAPVid-3D minival from $0.248$ to $0.255$
  (\S\ref{sec:eval}).
  \item \textbf{Highest absolute metric accuracy among methods evaluated under identical conditions}: combining the 3D refiner,
  the depth scale refiner, this second pool and the flow-only visibility head, our tracker attains the
  highest \emph{absolute} metric accuracy of all methods evaluated under identical conditions on a
  single commodity GPU: mean metric-Average Jaccard (AJ) $0.256$ (\S\ref{sec:eval}).
\end{enumerate}

We emphasise honest scoping throughout: our accuracy advantage is on the \emph{absolute} metric, where it
matters for downstream use; on the scale-invariant leaderboard a strong depth baseline and the
feed-forward tracker DELTA~\cite{delta} remain ahead, which we report explicitly.

\section{Related Work}
\label{sec:related}

\subsection{3D point tracking}
\label{sec:related-tracking}
TAPVid-3D~\cite{tapvid3d} formalised tracking any point in 3D and provides the minival split we use.
SpatialTracker~\cite{spatialtracker} and its successor SpatialTrackerV2~\cite{spatrackerv2} lift 2D
tracks into a learned 3D representation, the latter on a VGGT~\cite{vggt} visual-geometry front-end trained
end-to-end with the tracking head, but need a large temporal window to do so, which is memory-hungry;
TAPIP3D~\cite{tapip3d} aggregates tracks in a persistent world frame, which requires camera-motion
cancellation and hence \emph{precise} camera poses. DELTA~\cite{delta} is a per-frame feed-forward dense 3D tracker, and its
successor DELTAv2~\cite{deltav2} is lighter and reportedly faster at comparable accuracy;
TrackCraft3R~\cite{trackcraft3r} repurposes a heavy video-diffusion transformer for dense 3D tracking, at a
correspondingly high inference cost. These methods are strong on their intended settings, but several
depend on large windows, precise poses, or heavy backbones that do not fit a commodity single-GPU budget. We
report the resulting scores in \S\ref{sec:eval} under a fixed single-commodity-GPU, pose-free, monocular budget, where
only the feed-forward DELTA family remains competitive. Why the window- or
pose-dependent methods degrade is verified with each method's own official evaluator rather than attributed
to our harness.

\subsection{Optical flow and monocular depth}
\label{sec:related-flow-depth}
Our front-ends are frozen off-the-shelf models, chosen because each leads its benchmark at the time of
writing. SEA-RAFT~\cite{searaft} reports state-of-the-art accuracy on Spring~\cite{spring} and the best
cross-dataset generalization on KITTI~\cite{kitti} and Spring; WAFT~\cite{waft}, by the same authors, ranks
first on Spring, Sintel~\cite{sintel}, and KITTI while running faster than competitive alternatives. Both
provide dense optical flow for 2D correspondence.
Depth-Anything-3~\cite{da3} sets a new state of the art across camera pose, any-view geometry, and
monocular depth on its own visual-geometry benchmark, and provides monocular \emph{metric} depth for
the third dimension, in two backbone sizes (DA3-l, DA3-g) that we compare directly. As
an alternative depth source we also employ and evaluate
MegaSaM~\cite{megasam}, a deep visual-SLAM system that recovers dense \emph{metric} depth (together
with camera pose) from casual monocular video, using no external sensor. Both therefore take their metric
scale from a learned monocular prior---structure recovered from parallax alone is determined only up to
scale. They differ in what happens next, which is what matters under our absolute metric: MegaSaM refines
depth, pose and focal length in a joint bundle adjustment, and that solve is degenerate when the input
lacks parallax; its result can therefore drift far from the scale its prior supplied. We compare the two
front-ends under our absolute metric. We condition the refiner on frozen DINOv3~\cite{dinov3} features, whose
Gram-anchored, high-resolution dense patch descriptors give appearance context at the tracked pixel.

\subsection{State space models}
\label{sec:related-ssm}
Mamba~\cite{mamba} introduced selective state space models as a linear-time alternative to attention, but
its recurrent scan is causal, which is natural for language and a poor fit for images, where a token
should attend in every spatial direction. Three lines of work treat that causality as a constraint to
work around rather than remove: Vision Mamba~\cite{vim} runs separate forward and backward scans and
merges them, at roughly double the compute for the modest gain its own ablation reports;
MambaVision~\cite{mambavision} keeps the causal mixer and appends genuine self-attention blocks in its
final stages, abandoning the state space formulation exactly where global context matters; and
SF-Mamba~\cite{sfmamba} restructures the sequence to encode bidirectional flow under a still
unidirectional scan. Mamba-2~\cite{mamba2} established the \emph{state space duality} (SSD), an exact
identity between the recurrence and a masked form of linear attention, and VSSD~\cite{vssd} used it to
reinterpret the decay non-causally, which yields one genuinely non-causal operator instead of a
combination of scans or an appended fallback. Mamba-3~\cite{mamba3} separately improved the base
recurrence with an exponential-trapezoidal, three-term discretisation, but remains causal by
construction. Our refiner's mixer composes the last two of these, VSSD's non-causal collapse and Mamba-3's
three-term recurrence; the tracker needs only the temporal, single-direction case.

\section{Method}
\label{sec:method}
The tracker is built from two frozen front-ends and one small learned module.
\S\ref{sec:prelim} states the Mamba-3 layer that module uses; \S\ref{sec:method-track} onward builds the
pipeline around it.

\subsection{Preliminaries: Mamba-3 as masked linear attention}
\label{sec:prelim}
The only learned sequence operator in the pipeline is a Mamba-3~\cite{mamba3} layer applied along
each track's temporal axis. This subsection derives the two forms of it that the rest of
\S\ref{sec:method} refers to, Eqs.~\eqref{eq:ssd} and \eqref{eq:ncssd}. The operator's standalone
behaviour is not something the tracker relies on.

A selective state space model carries a hidden state $h_t\in\mathbb{R}^{N}$ through the linear
recurrence
\begin{equation}
  h_t = A_t\, h_{t-1} + B_t x_t, \qquad y_t = C_t^{\top} h_t,
  \label{eq:ssm}
\end{equation}
in the notation of~\cite{mamba3}, where the input $x_t\in\mathbb{R}$ and output
$y_t\in\mathbb{R}$ are scalars and $h_t,B_t,C_t\in\mathbb{R}^{N}$: the recurrence acts on one
channel at a time, and a layer runs $D$ of them in parallel.
Let $X\in\mathbb{R}^{T\times D}$ be a sequence of $T$ tokens of dimension $D$. Mamba-3 carries a hidden
state $h_t\in\mathbb{R}^{N}$ through an exponential-trapezoidal, three-term recurrence
\begin{equation}
  h_t = \alpha_t\, h_{t-1} + \beta_t\, B_{t-1} x_{t-1} + \gamma_t\, B_t x_t,
  \qquad y_t = C_t^{\top} h_t,
  \label{eq:rec3}
\end{equation}
with data-dependent scalars
\begin{equation}
  \alpha_t = e^{\Delta_t A_t},\qquad
  \beta_t  = (1-\lambda_t)\,\Delta_t\, e^{\Delta_t A_t},\qquad
  \gamma_t = \lambda_t\,\Delta_t,
  \label{eq:abc}
\end{equation}
where $\Delta_t$ is a step size, $A_t\in\mathbb{R}$ a state-transition scalar and $\lambda_t\in[0,1]$ a
learned convex weight, all produced from token $t$, and $B_t,C_t\in\mathbb{R}^{N}$ are its key and query
projections. Unrolling Eq.~\eqref{eq:rec3} and stacking over $t$ yields the \emph{state space duality} (SSD)
parallel form
\begin{equation}
  Y = \bigl(L \odot C B^{\top}\bigr)\, X,
  \label{eq:ssd}
\end{equation}
where $\odot$ is the Hadamard product and $L\in\mathbb{R}^{T\times T}$ is a structured mask, which
makes SSD a linear attention with no softmax, $L$ replacing it. That mask is
structured and lower-triangular. For Mamba-2, $L_{tj}=\gamma_j\prod_{k=j+1}^{t}\alpha_k$; the Mamba-3
recurrence adds a second band,
\begin{equation}
  L_{tj} = \Bigl(\textstyle\prod_{k=j+1}^{t}\alpha_k\Bigr)\gamma_j
         + \Bigl(\textstyle\prod_{k=j+2}^{t}\alpha_k\Bigr)\beta_{j+1},
  \qquad j\le t .
  \label{eq:L3}
\end{equation}
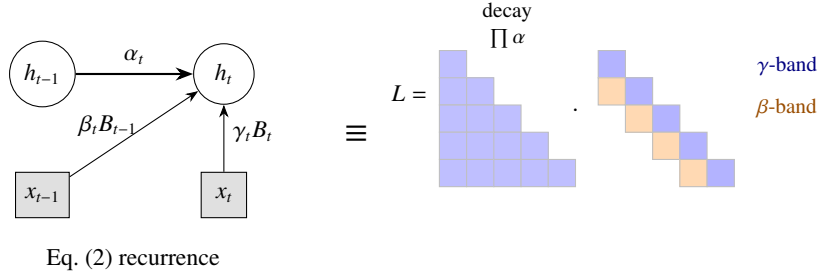
\begin{figure*}[!tb]
  \centering
  \begin{tikzpicture}[font=\small,>=Stealth]
    \node (hl) [circle,draw,minimum size=8mm] at (0,1.1) {$h_{t-1}$};
    \node (h)  [circle,draw,minimum size=8mm] at (2.4,1.1) {$h_{t}$};
    \draw[->,thick] (hl) -- node[above]{$\alpha_t$} (h);
    \node (xl) [rectangle,draw,fill=black!12,minimum size=6mm] at (0,-0.5) {$x_{t-1}$};
    \node (x)  [rectangle,draw,fill=black!12,minimum size=6mm] at (2.4,-0.5) {$x_{t}$};
    \draw[->] (xl) -- node[left,pos=0.6]{$\beta_t B_{t-1}$} (h);
    \draw[->] (x)  -- node[right,pos=0.6]{$\gamma_t B_t$} (h);
    \node at (1.2,-1.35) {Eq.~\eqref{eq:rec3} recurrence};
    \node (eq) at (4.15,0.3) {\Large$\equiv$};
    \node at (4.85,0.85) {$L=$};
    \begin{scope}[shift={(5.25,1.05)}]
      \foreach \r in {0,...,4}{
        \foreach \c in {0,...,\r}{
          \fill[blue!25] (\c*0.36,-\r*0.36) rectangle ++(0.36,0.36);
          \draw[gray!50] (\c*0.36,-\r*0.36) rectangle ++(0.36,0.36);
        }
      }
      \node[align=center] at (0.9,0.7) {\footnotesize decay\\[-1pt]\footnotesize $\textstyle\prod\alpha$};
    \end{scope}
    \node at (7.05,0.6) {$\cdot$};
    \begin{scope}[shift={(7.35,1.05)}]
      \foreach \r in {0,...,4}{
        \foreach \c in {0,...,\r}{
          \pgfmathtruncatemacro{\ondiag}{\r==\c ? 1 : 0}
          \pgfmathtruncatemacro{\onsub}{\r-\c==1 ? 1 : 0}
          \ifnum\ondiag=1
            \fill[blue!30] (\c*0.36,-\r*0.36) rectangle ++(0.36,0.36);
            \draw[gray!50] (\c*0.36,-\r*0.36) rectangle ++(0.36,0.36);
          \fi
          \ifnum\onsub=1
            \fill[orange!30] (\c*0.36,-\r*0.36) rectangle ++(0.36,0.36);
            \draw[gray!50] (\c*0.36,-\r*0.36) rectangle ++(0.36,0.36);
          \fi
        }
      }
      \node[blue!50!black] at (2.5,0.15) {\footnotesize $\gamma$-band};
      \node[orange!60!black] at (2.5,-0.4) {\footnotesize $\beta$-band};
    \end{scope}
    \node at (7.35,-2.05) {Eq.~\eqref{eq:ssd}: $Y=(L\odot CB^{\top})X$, with $L$ from Eq.~\eqref{eq:L3}};
  \end{tikzpicture}
  \caption{Mamba-3 state space duality. The three-term recurrence of Eq.~\eqref{eq:rec3} (left) is exactly a
  masked linear attention (right); ``$\equiv$'' denotes this duality---the two sides are the same
  computation, not merely similar. The structured mask $L$ is the product of a full lower-triangular decay
  matrix $\prod\alpha$ (every cell generically non-zero) and a two-band matrix carrying a main
  ($\gamma$) diagonal and, unique to Mamba-3, a sub-diagonal ($\beta$) band, Eq.~\eqref{eq:L3}---so $L$ itself is
  densely lower-triangular, not two isolated bands, and replaces the softmax of ordinary attention. Blank
  (upper-triangular) cells are zero in both matrices.}
  \label{fig:vm-ssd}
\end{figure*}
Reading Eq.~\eqref{eq:ssd} with $Q\!:=\!C$, $K\!:=\!B$, $V\!:=\!X$ exposes the duality: SSD \emph{is} linear
attention $Y=(L\odot QK^{\top})V$ with no softmax and no row normalisation, the structured mask $L$
replacing them and its decay $\prod_k\alpha_k$ acting as a relative-position bias
(Fig.~\ref{fig:vm-ssd}). This is the bridge the operators below exploit to build attention operators
with SSD's linear-time kernel.
Because $L$ is lower-triangular, Eq.~\eqref{eq:ssd} is causal. Reinterpreting $A_t$ as weighting the current
token rather than decaying the past makes the hidden state independent of the query index, and the
$T\times T$ mask collapses to a single $T$-vector $m$ with $m_j=1/A_j$~\cite{vssd}:
\begin{equation}
  Y = C\bigl((B\odot m)^{\top} X\bigr),
  \label{eq:ncssd}
\end{equation}
at $O(TND)$ cost with an $O(ND)$ state independent of the sequence length $T$. That constant-state
property is why the refiner is built on Mamba-3 rather than softmax attention: a track can span hundreds
of frames, and Eq.~\eqref{eq:ncssd} processes them at a fixed per-step memory budget. We write this operator
VSSD-1pool, after the single pool its vector reads, to distinguish it from the two-pool
VSSD-2pool of \ref{sec:vmamba3-noncausal}, which explains why the naming counts pools rather than
the bands of $L$. Equation~\eqref{eq:ncssd} is the form both
refiners and the de-flicker stage use along time.
\subsection{Non-causal Mamba-3 self- and cross-attention}
\label{sec:vmamba3}
We abbreviate the construction below \emph{vmamba3} for readability. Nothing in it is claimed as
novel:
Mamba-3~\cite{mamba3} is not ours, nor is the non-causal collapse~\cite{vssd} or the multi-directional
scanning~\cite{vmamba} this section builds on, and \emph{Vision Mamba}~\cite{vim} already names a
separate line of work. The vmamba3 refiner of \S\ref{sec:track-vmamba3-refiner}, for instance,
adds no positional encoding of its own: its inputs already carry pre-encoded spatial context, and it
simply applies Mamba-3 self-attention to those features.

\subsubsection{Self- and cross-attention}
\label{sec:vmamba3-selfcross}
Feeding a value projection $V=\mathrm{SiLU}(XW_V)$ into Eq.~\eqref{eq:ssd} gives a self-attention layer
\begin{equation}
  Y^{\mathrm{self}} = \bigl(L\odot CB^{\top}\bigr)V,
  \label{eq:self}
\end{equation}
causal because $L_{ij}=0$ for $j>i$. Cross-attention lets the query
side supply $C$ while a key/value side supplies $B$ and $V$, either in a state-compressed form with
constant memory in the key length, or in an exact token-level form
\begin{equation}
  Y^{\mathrm{cross}} = \bigl(L^{\mathrm{cross}}\odot C^{q}B^{kv\top}\bigr)V^{kv}.
  \label{eq:cross}
\end{equation}

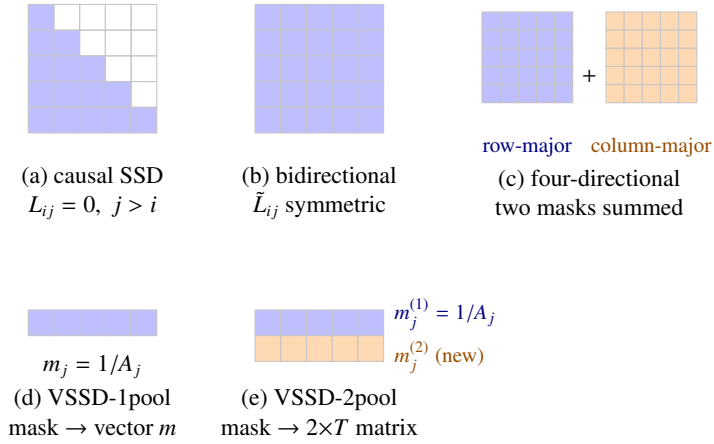
\begin{figure*}[!tb]
  \centering
  \begin{tikzpicture}[font=\small]
    \begin{scope}[shift={(0,0)}]
      \foreach \r in {0,...,4}\foreach \c in {0,...,4}{
        \pgfmathtruncatemacro{\on}{\c<=\r ? 1 : 0}
        \ifnum\on=1 \fill[blue!25] (\c*0.34,-\r*0.34) rectangle ++(0.34,0.34);\fi
        \draw[gray!45] (\c*0.34,-\r*0.34) rectangle ++(0.34,0.34);
      }
      \node[align=center] at (0.85,-2.15) {(a) causal SSD\\ $L_{ij}=0,\ j>i$};
    \end{scope}
    \begin{scope}[shift={(3.0,0)}]
      \foreach \r in {0,...,4}\foreach \c in {0,...,4}{
        \fill[blue!25] (\c*0.34,-\r*0.34) rectangle ++(0.34,0.34);
        \draw[gray!45] (\c*0.34,-\r*0.34) rectangle ++(0.34,0.34);
      }
      \node[align=center] at (0.85,-2.15) {(b) bidirectional\\ $\tilde L_{ij}$ symmetric};
    \end{scope}
    \begin{scope}[shift={(6.0,0)}]
      \foreach \r in {0,...,4}\foreach \c in {0,...,4}{
        \fill[blue!25] (\c*0.24,-\r*0.24) rectangle ++(0.24,0.24);
        \draw[gray!45] (\c*0.24,-\r*0.24) rectangle ++(0.24,0.24);
      }
      \node at (1.42,-0.6) {$+$};
      \begin{scope}[shift={(1.64,0)}]
        \foreach \r in {0,...,4}\foreach \c in {0,...,4}{
          \fill[orange!30] (\c*0.24,-\r*0.24) rectangle ++(0.24,0.24);
          \draw[gray!45] (\c*0.24,-\r*0.24) rectangle ++(0.24,0.24);
        }
      \end{scope}
      \node[align=center,blue!50!black] at (0.6,-1.55) {\footnotesize row-major};
      \node[align=center,orange!60!black] at (2.24,-1.55) {\footnotesize column-major};
      \node[align=center] at (1.42,-2.15) {(c) four-directional\\ two masks summed};
    \end{scope}
    \begin{scope}[shift={(0,-3.7)}]
      \foreach \c in {0,...,4}{
        \fill[blue!25] (\c*0.34,0) rectangle ++(0.34,-0.34);
        \draw[gray!45] (\c*0.34,0) rectangle ++(0.34,-0.34);
      }
      \node[align=center] at (0.85,-1.35) {(d) VSSD-1pool\\ mask $\to$ vector $m$};
      \node at (0.85,-0.75) {$m_j=1/A_j$};
    \end{scope}
    \begin{scope}[shift={(3.0,-3.7)}]
      \foreach \c in {0,...,4}{
        \fill[blue!25] (\c*0.34,0) rectangle ++(0.34,-0.34);
        \draw[gray!45] (\c*0.34,0) rectangle ++(0.34,-0.34);
        \fill[orange!30] (\c*0.34,-0.34) rectangle ++(0.34,-0.34);
        \draw[gray!45] (\c*0.34,-0.34) rectangle ++(0.34,-0.34);
      }
      \node[right,font=\footnotesize,blue!50!black] at (1.72,-0.06) {$m^{(1)}_j=1/A_j$};
      \node[right,font=\footnotesize,orange!60!black] at (1.72,-0.62) {$m^{(2)}_j$ (new)};
      \node[align=center] at (0.85,-1.35) {(e) VSSD-2pool\\ mask $\to$ $2{\times}T$ matrix};
    \end{scope}
  \end{tikzpicture}
  \caption{The causal mask, and the four ways of \ref{sec:vmamba3-noncausal} to remove causality from Eq.~\eqref{eq:ssd}. (a) Causal SSD masks the future. (b) Summing
  forward and reverse scans yields a symmetric, all-pairs mask---non-causal, but still weighted by distance
  along the scan order (\ref{sec:vmamba3-noncausal}). (c) Four-directional scanning sums a second,
  column-major symmetric mask on top of (b)'s row-major one, removing the direction bias a single scan
  order leaves behind. (d) VSSD-1pool~\cite{vssd} collapses the $T\times T$ mask to a single $T$-vector
  $m$, so Eq.~\eqref{eq:ncssd} runs in $O(TND)$ with an $O(ND)$ state independent of sequence length, but
  cannot represent Mamba-3's $\beta$-band. (e) VSSD-2pool stacks a second, independently-read
  vector $m^{(2)}$ under $m^{(1)}$, recovering $\beta$-band-like capacity at $2\times$ VSSD-1pool's cost,
  Eq.~\eqref{eq:vssd-2pool}; derived in \ref{sec:vmamba3-noncausal} and measured in Table~\ref{tab:cifar}.
  Coloured cells denote non-zero mask entries; blank cells are zero (masked out). Where a panel sums two contributions, the second is drawn in orange---the same orange Fig.~\ref{fig:vm-ssd} uses for the $\beta$-band, which in (e) is literally what $m^{(2)}$ is motivated by.}
  \label{fig:vm-attn}
\end{figure*}

\subsubsection{Removing causality}
\label{sec:vmamba3-noncausal}
A raster-flattened image token that only attends to earlier tokens is blind to half its spatial
neighbourhood; the causal mask of Eq.~\eqref{eq:L3} is inappropriate for 2D data. Causality can be
removed either by scanning the sequence in several directions, which is what a raster-flattened
\emph{image} needs, or by collapsing the mask to a per-token vector and holding a state whose size does
not grow with the sequence. Figure~\ref{fig:vm-attn} draws the causal mask and the four ways we
considered of removing causality from it: two scan constructions and two collapses. The tracker mixes
along time only and therefore uses a collapse; the four are set out and measured against one another
in \ref{sec:noncausal-constructions}.

The collapse that already exists is VSSD~\cite{vssd}, which we write \emph{VSSD-1pool}: the
query-independent form of Eq.~\eqref{eq:ncssd}, whose $T\times T$ mask reduces to a single $T$-vector
$m$ with $m_j=1/A_j$ (Fig.~\ref{fig:vm-attn}(d)). It cannot represent a relative, $|i-j|$-dependent
bias, since $m_j$ is independent of $i$, so Mamba-3's extra $\beta$-band of Eq.~\eqref{eq:L3} does not
survive it: VSSD-1pool applied to Mamba-3's projections is structurally a Mamba-2-style reduction,
trading per-pair adaptivity for an $O(ND)$ state. Recovering that band is what the operator below adds.

\emph{VSSD-2pool}: the causal $\beta$-band is in fact recoverable in $m$ at no extra cost by
redefining $m_j:=\gamma_j+\beta_{j+1}$, because both bands then sit on the same side of the diagonal.
The non-causal collapse of VSSD-1pool is not: its single vector serves queries on both sides at once, where
the $\beta$ contribution comes from $\beta_{j+1}$ looking back and from $\beta_{j-1}$ looking forward,
and one $m_j$ --- independent of the query index by construction --- cannot hold both. The fix is a second, independently-read pool: its own query projection
$C^{(2)}$ paired with a second vector $m^{(2)}$, stacking $m$ into a $2\times T$ matrix
(Fig.~\ref{fig:vm-attn}(e)):
\begin{equation}
  Y = C^{(1)}\bigl((B\odot m^{(1)})^{\top} X\bigr) + C^{(2)}\bigl((B\odot m^{(2)})^{\top} X\bigr),
  \label{eq:vssd-2pool}
\end{equation}
with $m^{(1)}_j=1/A_j$ unchanged and $m^{(2)}_j$ a second, freely-learned per-token vector, at $2\times$
VSSD-1pool's state and compute. This is a motivated capacity extension, not a proof that the second pool
reproduces Mamba-3's causal $\beta$-band behaviour.

VSSD-2pool is our default: it is the more accurate of the two collapses in the tracker, raising mean
absolute metric-AJ from $0.248$ to $0.255$, and every \texttt{vmamba3-2pool} arm uses it. The depth
scale refiner is the one exception and uses VSSD-1pool.

\subsection{Proposed method overview}
\label{sec:method-track}
Given a monocular video and query points, we produce a metric 3D track per query in three stages
(Fig.~\ref{fig:pipeline}): (i) propagate each query in 2D with frozen optical flow; (ii) lift the 2D track
to metric 3D with frozen monocular depth; (iii) refine only the per-point depth along its fixed pixel ray
with a compact Mamba-3 module. Only stage (iii) is learned.

\subsection{2D track (frozen)}
\label{sec:track-2d}
We precompute dense flow between consecutive frames, forward and backward, with
SEA-RAFT~\cite{searaft} or the stronger WAFT~\cite{waft}. One such field moves a point by a single
frame; a full-length trajectory therefore comes from \emph{flow chaining} (the box of that name in
Fig.~\ref{fig:pipeline}): each query starts at its anchor frame, and we place it in the next frame by
sampling the next flow field at its current sub-pixel position, bilinearly blending the displacements of
the four surrounding pixels, then repeat from
that new position---sweeping forward to later frames and backward to earlier ones. Sampling at the
propagated position rather than at the query pixel is what accumulates the hops into a long-range
track. Visibility follows a forward--backward consistency test: propagating a point forward and then
back through the same flow field should return it close to where it started; a large round-trip
error means the flow there could not be trusted, which is exactly what happens under occlusion,
since flow assumes a one-to-one pixel correspondence that an occluded point does not have. A point
that fails this test stays occluded.

\subsection{Metric depth and unprojection}
\label{sec:track-depth}
Depth-Anything-3~\cite{da3} predicts a per-frame metric depth map $D_t$. We use two frozen instances of
it that differ by $4\times$ in parameter count: \emph{DA3-l} (\texttt{DA3Metric-Large}, $0.35$\,B), a
single distilled monocular network, and \emph{DA3-g} (\texttt{DA3Nested-Giant-Large}, $1.40$\,B), which
nests a multi-view geometry branch with the DA3-l branch and fuses them by a per-frame least-squares
scale fit. The two are metrically wrong in different ways---DA3-l carries a far-field bias, DA3-g a
frame-to-frame scale flicker, and we report them separately; unless noted,
DA3-l is our default and headline backbone. Sampling either at the tracked pixel gives a raw depth
$z^{\mathrm{raw}}_t=D_t(\mathbf{u}_t)$, and the pinhole model unprojects the point to
camera-frame 3D:
\begin{equation}
  \mathbf{X}_t = z_t \,\Bigl(\tfrac{u_t-c_x}{f_x},\ \tfrac{v_t-c_y}{f_y},\ 1\Bigr)^{\!\top},
  \label{eq:unproject}
\end{equation}
with intrinsics $(f_x,f_y,c_x,c_y)$. Taking $z_t=z^{\mathrm{raw}}_t$ is our training-free baseline.

\subsection{Two 3D refiner variants}
\label{sec:track-variants}
On top of this baseline we develop \emph{two} learned 3D refiners of increasing capacity, and we compare
them directly in \S\ref{sec:eval}. The \emph{Mamba-3 3D refiner} (depth-along-ray) corrects only the depth
of each point along its fixed pixel ray, from purely scale-invariant geometric cues. The
\emph{vmamba3 3D refiner} (appearance-conditioned) additionally conditions on frozen appearance and
local-geometry features and is allowed a small 2D correction of the tracked pixel. Both run Mamba-3
over the per-track temporal axis and differ only in their inputs and in whether the 2D position is
refined; we describe them in turn.

Both refiners are \emph{non-causal in time}. Their temporal mixing uses the query-independent collapse of
Eq.~\eqref{eq:ncssd} rather than a causal mask, and the correction at frame $t$ is therefore computed from one pooled
summary of the whole clip: later frames inform it exactly as much as earlier ones. The consequence is
that the refiners work at clip level and not online---they need the whole clip in hand before producing
anything, and a streaming tracker would instead require the causal mask we deliberately do not apply.
That suits this application, where depth is read from an offline cache and the per-clip median
$z_{\mathrm{ref}}$ below is itself a clip-level statistic.

Neither refiner mixes over \emph{space}. Both use the temporal collapse of Eq.~\eqref{eq:ncssd}, and neither
needs the multi-directional scans (i)--(ii) of \ref{sec:noncausal-constructions}, which exist to give a
raster-flattened \emph{image} a non-causal neighbourhood, nor any of the three positional encodings of
\ref{sec:vmamba3-pos}, 2-D RoPE included. The reason is that each refiner's
two extra inputs already carry spatial context from elsewhere---the DINOv3 feature is a whole-image summary at
the tracked pixel, produced by DINOv3's own frozen self-attention, and the local depth patch's $25$ samples
occupy fixed grid positions an MLP can learn to interpret directly---the refiner therefore only has to mix that
context along time.

\subsection{Mamba-3 3D refiner: depth-along-ray correction}
\label{sec:track-mamba3-refiner}
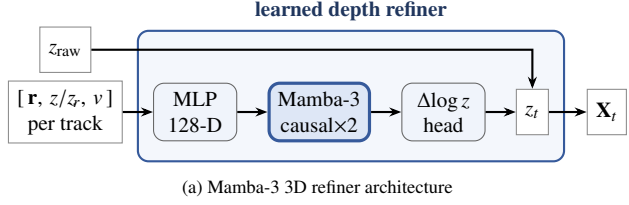
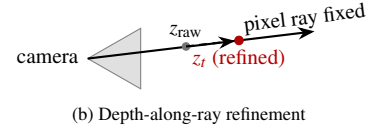
\begin{figure*}[!tb]
\centering
\begin{subfigure}[b]{0.62\linewidth}
\centering
\begin{tikzpicture}[
    node distance=3mm and 4mm,
    block/.style={rectangle, rounded corners, draw=black!60, fill=cTrain!8,
      minimum height=7mm, minimum width=11mm, align=center, font=\footnotesize},
    frozen/.style={rectangle, rounded corners, draw=black!50, fill=black!12,
      minimum height=7mm, minimum width=11mm, align=center, font=\footnotesize},
    learn/.style={rectangle, rounded corners, draw=cTrain!75!black, fill=cTrain!18, very thick,
      minimum height=7mm, minimum width=13mm, align=center, font=\footnotesize},
    io/.style={rectangle, draw=black!45, fill=white,
      minimum height=6mm, align=center, font=\footnotesize},
    arr/.style={-{Stealth[length=1.6mm]}, thick},
  ]
  \node[io] (in) {$[\,\mathbf{r},\,z/z_{\!r},\,v\,]$\\per track};
  \node[block, right=of in] (emb) {MLP\\128-D};
  \node[learn, right=of emb] (m1) {Mamba-3\\causal$\times$2};
  \node[block, right=of m1] (head) {$\Delta\!\log z$\\head};
  \node[io, right=of head] (z) {$z_t$};
  \node[io, right=5mm of z] (out) {$\mathbf{X}_t$};
  \node[io, above=1mm of in] (zraw) {$z_{\text{raw}}$};
  \draw[arr] (in) -- (emb);
  \draw[arr] (emb) -- (m1);
  \draw[arr] (m1) -- (head);
  \draw[arr] (head) -- (z);
  \draw[arr] (z) -- (out);
  \draw[arr] (zraw.east) -| (z.north);
  \coordinate (zrband) at (zraw.east -| z);
  \begin{scope}[on background layer]
    \node[draw=cTrain!75!black, thick, rounded corners, fill=cTrain!6,
          fit=(emb)(m1)(head)(z)(zrband), inner xsep=2mm, inner ysep=2.5mm,
          label={[font=\footnotesize\bfseries, cTrain!45!black]above:{learned depth refiner}}] {};
  \end{scope}
\end{tikzpicture}
\subcaption{Mamba-3 3D refiner architecture}
\label{fig:arch}
\end{subfigure}%
\hfill
\begin{subfigure}[b]{0.34\linewidth}
\centering
\begin{tikzpicture}[scale=1.0]
  \coordinate (cam) at (0,0);
  \draw[fill=black!12,draw=black!50] (cam) -- (0.7,0.40) -- (0.7,-0.40) -- cycle;
  \node[font=\footnotesize,left] at (cam) {camera};
  \draw[arr,black,thick] (cam) -- (3.0,0.36)
    node[pos=0.68,sloped,anchor=south west,inner sep=1pt,font=\footnotesize,text=black]{pixel ray fixed};
  \filldraw[black!55] (1.3,0.16) circle (1.4pt)
    node[above,font=\footnotesize,text=black]{$z_{\text{raw}}$};
  \filldraw[red!70!black] (2.0,0.24) circle (1.6pt)
    node[below,font=\footnotesize,text=red!70!black]{$z_t$ (refined)};
  \draw[arr,densely dashed] (1.3,0.16) -- (1.95,0.235);
\end{tikzpicture}
\subcaption{Depth-along-ray refinement}
\label{fig:ray}
\end{subfigure}
\caption{Mamba-3 3D refiner.
\textbf{(a)} Architecture --- the inside of the blue box in Fig.~\ref{fig:pipeline}.
$\mathbf{r}{=}(\text{ray}_x,\text{ray}_y)$ is the frozen pixel ray from the 2D track $\mathbf{u}_t$;
$z/z_{\!r}{=}z_{\text{raw}}/z_{\text{ref}}$; $v{=}$visibility. \emph{Bypass}: $z_{\text{raw}}$ (the frozen
metric-depth backbone's raw depth) skips the SSM entirely and is multiplied back in via
$z_t=z_{\text{raw}}e^{\Delta\log z}$, Eq.~\eqref{eq:correction}; the SSM is therefore a \emph{relative
corrector}, not an absolute-depth predictor. \emph{causal$\times$2}: two causal Mamba-3 layers
(past-only; enables streaming). SSD splits the 64-dim state into 4 heads (16-dim each), analogous to
multi-head attention but computed as a linear recurrence.
\textbf{(b)} Ray-invariance guarantee. \emph{Red marks what the refiner changes, black what it leaves
alone} --- the same convention as Fig.~\subfigref{fig:v35-arch-concept}{fig:v35-concept}. The grey wedge is the camera:
apex at the camera centre, right edge the image plane; the ray leaves the centre through the tracked
pixel and continues into the scene. The pixel, and hence the ray direction, is frozen, so \emph{only
the depth slides along the ray} and the 2D projection is unchanged: the refiner moves depth alone,
leaving the 2D track and visibility invariant.}
\label{fig:arch-ray}
\end{figure*}

With the pixel frozen, the only degree of freedom that preserves the 2D reprojection is the depth $z_t$
along the ray. The Mamba-3 refiner (Fig.~\subfigref{fig:arch-ray}{fig:arch}) therefore predicts a
multiplicative correction
\begin{equation}
  z_t = z^{\mathrm{raw}}_t \, e^{\Delta \log z_t},
  \label{eq:correction}
\end{equation}

in log-depth rather than raw depth: a fixed relative depth error grows without bound in absolute meters as
range increases, and no single additive correction $\Delta z_t$ can therefore fit both a near and a far point equally
well, whereas a multiplicative correction scales with $z^{\mathrm{raw}}_t$ automatically; DELTA~\cite{delta}
independently reports log-depth as the empirically best depth representation for 3D tracking. The Mamba-3
refiner takes scale-invariant inputs---the normalised ray direction, the depth ratio $z^{\mathrm{raw}}_t/z_{\mathrm
{ref}}$ (with $z_{\mathrm{ref}}$ the per-clip median depth), and the visibility flag---embedded by an MLP
and processed by two Mamba-3 layers over the temporal axis, with a zero-initialised
$\Delta\log z$ head so that the refiner starts exactly at the baseline.

\subsection{vmamba3 3D refiner: appearance- and geometry-conditioned correction}
\label{sec:track-vmamba3-refiner}
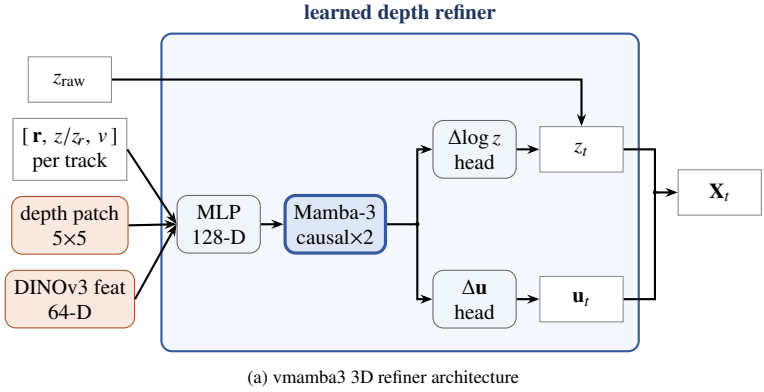
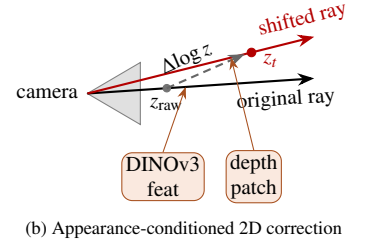
\begin{figure*}[!tb]
\centering
\begin{subfigure}[b]{0.64\linewidth}
\centering
\begin{tikzpicture}[
    node distance=2.5mm and 3mm,
    block/.style={rectangle, rounded corners, draw=black!60, fill=cTrain!8,
      minimum height=6.5mm, minimum width=11mm, align=center, font=\footnotesize},
    learn/.style={rectangle, rounded corners, draw=cTrain!75!black, fill=cTrain!18, very thick,
      minimum height=6.5mm, minimum width=13mm, align=center, font=\footnotesize},
    io/.style={rectangle, draw=black!45, fill=white,
      minimum height=6mm, minimum width=11mm, align=center, font=\footnotesize},
    new/.style={rectangle, rounded corners, draw=cInput!75!black, fill=cInput!15,
      minimum height=6.5mm, minimum width=14mm, align=center, font=\footnotesize},
    arr/.style={-{Stealth[length=1.6mm]}, thick},
  ]
  \node[io] (in1) {$[\,\mathbf{r},\,z/z_{\!r},\,v\,]$\\per track};
  \node[new, below=2mm of in1] (patch) {depth patch\\$5{\times}5$};
  \node[new, below=2mm of patch] (dino) {DINOv3 feat\\64-D};
  \coordinate (embw) at ($(in1.east)!0.5!(dino.east)$);
  \node[block, anchor=west] (emb) at ([xshift=6mm]embw) {MLP\\128-D};
  \node[learn, right=of emb] (ssm) {Mamba-3\\causal$\times$2};
  \node[block, above right=2mm and 6mm of ssm] (dz)  {$\Delta\!\log z$\\head};
  \node[block, below right=2mm and 6mm of ssm] (duv) {$\Delta\mathbf{u}$\\head};
  \node[io, right=3mm of dz]  (zout)  {$z_t$};
  \node[io, right=3mm of duv] (uvout) {$\mathbf{u}_t$};
  \node[io, above=2mm of in1] (dm) {$z_{\text{raw}}$};
  \draw[arr] (in1.east)   -- (emb.west);
  \draw[arr] (patch.east) -- (emb.west);
  \draw[arr] (dino.east)  -- (emb.west);
  \draw[arr] (emb) -- (ssm);
  \coordinate (ssmfork) at ($(ssm.east)+(4mm,0)$);
  \draw[thick] (ssm.east) -- (ssmfork);
  \fill (ssmfork) circle[radius=0.8pt];
  \draw[arr] (ssmfork) |- (dz.west);
  \draw[arr] (ssmfork) |- (duv.west);
  \draw[arr] (dz)  -- (zout);
  \draw[arr] (duv) -- (uvout);
  \draw[arr] (dm.east) -| (zout.north);
  \coordinate (boxtop) at (dz.center |- dm.north);
  \begin{scope}[on background layer]
    \node[draw=cTrain!75!black, thick, rounded corners, fill=cTrain!6,
          fit=(emb)(ssm)(dz)(duv)(zout)(uvout)(boxtop), inner xsep=2mm, inner ysep=3mm,
          label={[font=\footnotesize\bfseries, cTrain!45!black]above:{learned depth refiner}}] (refbox) {};
  \end{scope}
  \node[io, right=5mm of refbox.east] (out) {$\mathbf{X}_t$};
  \coordinate (omerge) at ($(out.west)+(-3mm,0)$);
  \draw[thick] (zout.east)  -| (omerge);
  \draw[thick] (uvout.east) -| (omerge);
  \fill (omerge) circle[radius=0.8pt];
  \draw[arr] (omerge) -- (out.west);
\end{tikzpicture}
\subcaption{vmamba3 3D refiner architecture}
\label{fig:v35-arch}
\end{subfigure}%
\hfill
\begin{subfigure}[b]{0.32\linewidth}
\centering
\begin{tikzpicture}[scale=1.0]
  \coordinate (cam) at (0,0);
  \draw[fill=black!12,draw=black!50] (cam) -- (0.7,0.41) -- (0.7,-0.41) -- cycle;
  \node[font=\footnotesize,left] at (cam) {camera};
  \draw[arr,black,thick] (cam) -- (3.0,0.20)
    node[pos=0.87,below,sloped,font=\footnotesize,text=black]{original ray};
  \filldraw[black!55] (1.05,0.070) circle (1.4pt)
    node[below,font=\footnotesize,text=black]{$z_{\rm raw}$};
  \draw[arr,red!70!black,thick] (cam) -- (3.0,0.75)
    node[pos=0.97,above,sloped,font=\footnotesize,red!70!black]{shifted ray};
  \filldraw[red!70!black] (2.17,0.5425) circle (1.6pt)
    node[right,xshift=1pt,yshift=-3pt,font=\footnotesize,text=red!70!black]{$z_t$};
  \draw[arr,densely dashed,black!60] (1.05,0.070) -- (2.08,0.52)
    node[pos=0.35,sloped,above,font=\footnotesize,text=black]{$\Delta\!\log z$};
  \node[draw=cInput!75!black,fill=cInput!15,rounded corners,font=\footnotesize,inner sep=1.5pt,align=center]
    (dinobox) at (1.0,-1.1) {DINOv3\\feat};
  \draw[-{Stealth[length=1.6mm]},cInput!75!black] (dinobox.north) -- (1.3,0.087);
  \node[draw=cInput!75!black,fill=cInput!15,rounded corners,font=\footnotesize,inner sep=1.5pt,align=center]
    (patchbox) at (2.2,-1.1) {depth\\patch};
  \draw[-{Stealth[length=1.6mm]},cInput!75!black] (patchbox.north) -- (1.9,0.475);
\end{tikzpicture}
\subcaption{Appearance-conditioned 2D correction}
\label{fig:v35-concept}
\end{subfigure}
\caption{vmamba3 3D refiner.
\textbf{(a)} Architecture, extending Fig.~\subfigref{fig:arch-ray}{fig:arch} --- again the inside of the blue box in
Fig.~\ref{fig:pipeline}. Orange nodes are new relative to the Mamba-3 refiner. The pixel position
$\mathbf{u}_0$ (from the frozen optical-flow front-end; Fig.~\ref{fig:pipeline}) sets the ray direction
$\mathbf{r}(\mathbf{u}_0)$, the depth-patch centre, and the DINOv3 sample location. The three inputs are
embedded and fed through the same causal Mamba-3 SSD stack, now carrying appearance and local-geometry
context. A second, zero-initialised head predicts $\Delta\mathbf{u}=2\tanh(\cdot)$ (bounded to
$\pm2$\,px); depth is re-sampled at $\mathbf{u}_t=\mathbf{u}_0+\Delta\mathbf{u}$ before $\Delta\log z$
rescales it. Both heads start at zero, and the vmamba3 refiner therefore starts exactly at the Mamba-3
refiner's output.
\textbf{(b)} Safe, appearance-conditioned 2D correction. \emph{Red marks what changes, black what does
not}: the shifted ray and $z_t$ are both red because this variant moves both, whereas
Fig.~\subfigref{fig:arch-ray}{fig:ray} moves only $z_t$. The camera wedge is as in Fig.~\subfigref{fig:arch-ray}{fig:ray}; both rays
leave the centre through their own pixel. Two new inputs --- DINOv3 appearance and a local
$5{\times}5$ depth patch, both orange --- give the model the context to predict a 2D shift
$\Delta\mathbf{u}$ safely: unlike a naive nudge, which crosses depth discontinuities, this correction is
conditioned on local appearance and geometry. Depth is then re-sampled at the shifted pixel
$\mathbf{u}_0{+}\Delta\mathbf{u}$ before $\Delta\log z$ rescales it.}
\label{fig:v35-arch-concept}
\end{figure*}

A \emph{naive} 2D correction---letting the refiner nudge the tracked pixel from geometry alone---in fact
\emph{degraded} accuracy in our experiments: the frozen flow track is already near-optimal, and a blind
nudge tends to slide the point across a depth discontinuity onto the wrong surface. The lesson is that
depth alone cannot tell the refiner \emph{which} surface it is looking at; the 2D correction therefore needs
appearance context to move safely. The vmamba3 refiner (Fig.~\subfigref{fig:v35-arch-concept}{fig:v35-arch}) therefore augments
the Mamba-3 refiner---which otherwise leaves the 2D track untouched throughout, only the depth sliding along
its frozen ray (Fig.~\subfigref{fig:arch-ray}{fig:ray})---with two per-track inputs: a $64$-D DINOv3~\cite{dinov3} appearance
feature sampled at the tracked pixel from a frozen ViT, and a $5\times5$ local depth patch ($25$ bilinear
depth samples on a grid around the pixel, each divided by the centre depth to stay scale-invariant).

Together
these let a second, zero-initialised head predict a small bounded 2D correction
$\Delta\mathbf{u}=2\tanh(\cdot)$ (at most $\pm2$\,px) safely, without crossing a depth discontinuity
(Fig.~\subfigref{fig:v35-arch-concept}{fig:v35-concept}); depth is then re-sampled at the corrected pixel before Eq.~\eqref{eq:correction}.
The full refiner has $0.44$\,M trainable parameters with one pool and $0.61$\,M with two; the frozen flow, depth, and DINOv3 networks carry
all the heavy representation. Both refiners mix along time with VSSD-1pool; the
\texttt{vmamba3-2pool} arms swap it for VSSD-2pool, changing nothing else.

\subsection{Depth scale refiner: stabilising a frame-to-frame scale before refinement}
\label{sec:scale-refiner}
Our larger metric-depth backbone (DA3-g, \S\ref{sec:track-depth}) is individually more accurate at
range, but its fused metric scale is re-solved independently every frame, and therefore drifts frame
to frame even where the underlying depth shape is smooth.
Neither 3D refiner above can see this: both mix information only along \emph{time}, one track at a
time, and neither therefore observes a quantity shared by every point in a frame --- and a
per-frame, all-points-shared scalar is exactly what a frame-local flicker is. We therefore add a
small stage \emph{before} the refiner rather than trying to correct it after.

\emph{De-flicker, the first form of that stage.} Per frame it mean-pools the visible tracked points
--- each described by its ray direction, its depth relative to the clip's anchor-frame median, and
its visibility --- into a single frame token, mixes those tokens along time, and emits one
log-scale correction $\Delta s_f$ per frame, applied to depth along the frozen ray. The head is
zero-initialised; the stage therefore begins exactly at the un-corrected baseline, and any gain it
shows is its own. \S\ref{sec:eval} reports it both standalone and followed by the 3D refiner, the
\emph{de-flicker-then-refine} arms.

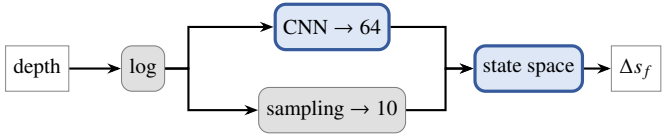
\begin{figure}[!tb]
\centering
\begin{tikzpicture}[font=\footnotesize,
  bx/.style={draw, minimum height=6mm, inner xsep=3pt, align=center},
  ob/.style={bx, sharp corners, fill=white, draw=black!45},
  pr/.style={bx, rounded corners, fill=black!12, draw=black!50},
  lr/.style={bx, rounded corners, fill=cTrain!18, draw=cTrain!75!black, very thick},
  a/.style={-{Stealth[length=1.8mm]}, thick}]
  \node[ob] (d) at (0,0)      {depth};
  \node[pr] (l) at (1.4,0)    {log};
  \node[lr] (c) at (3.9,0.55) {CNN $\rightarrow 64$};
  \node[pr] (q) at (3.9,-0.55){sampling $\rightarrow 10$};
  \node[lr] (m) at (6.5,0)    {state space};
  \node[ob] (o) at (7.92,0)    {$\Delta s_f$};
  \draw[a] (d) -- (l);
  \coordinate (x) at (2.0,0);
  \draw[a] (l.east) -- (x) |- (c.west);
  \draw[a] (l.east) -- (x) |- (q.west);
  \coordinate (y) at (5.4,0);
  \draw[a] (c.east) -| (y) -- (m.west);
  \draw[a] (q.east) -| (y) -- (m.west);
  \draw[a] (m) -- (o);
\end{tikzpicture}
\caption{\textbf{Depth scale refiner} (\S\ref{sec:scale-refiner}). It reads the metric depth
  map directly: no tracker appears in the path, and the front-end cannot change what it learns.
  Two independent branches summarise each frame's log depth: a small CNN over it pooled to $64^2$,
  giving $64$ features, and the $5$th, $15$th, \ldots, $95$th percentiles taken over \emph{every}
  pixel, one vector of ten numbers per frame. Pooling and then averaging inside the CNN discards
  the shape of the depth distribution, which the percentiles restore. The two concatenate to $74$ and project to one
  $128$-dimensional token per frame; the state space layer mixes the $F$ tokens and a linear
  head emits one scalar $\Delta s_f$, rescaling that frame's depth
  $z\leftarrow z^{\mathrm{raw}}e^{\Delta s_f}$ before the refiner of
  \S\ref{sec:track-vmamba3-refiner} runs.}
\label{fig:scale-refiner}
\end{figure}

\emph{What the stage must not depend on.} De-flicker reads the drift \emph{through the tracker}:
depth sampled at flow-chosen pixels, averaged under a flow-derived visibility mask. The drift is a
property of the depth model alone, and this therefore couples the stage to a front-end that has
nothing to do with the quantity being estimated, and two copies trained behind different flow networks then learn
different corrections from identical depth. The \emph{depth scale refiner} keeps the same
per-frame, zero-initialised, time-mixed form and removes that coupling: it reads the depth map
directly and never sees a tracked point (Fig.~\ref{fig:scale-refiner}).

Per frame it forms two independent summaries of the log depth: a small convolutional encoder over
the map pooled to $64^2$, giving $64$ features, and ten percentiles of the log depth taken over
every pixel --- the $5$th, $15$th, \ldots, $95$th. The second exists because the encoder ends in a
global average, which discards the shape of the depth distribution; the percentiles put that shape
back. The two concatenate to $74$ and project to one $128$-dimensional token per frame, and a
time-axis layer mixes those tokens across the whole clip, symmetrically in time: rescaling a cached
clip is free to look ahead. A zero-initialised head then
predicts a single per-frame log-scale correction $\Delta s_f$; every point in that frame is rescaled
$z\leftarrow z^{\mathrm{raw}}\,e^{\Delta s_f}$ before unprojection. The vmamba3 3D refiner then runs on
this temporally stabilised depth, and its along-ray and 2D corrections work with a consistent signal
instead of fighting the drift.

\subsection{Visibility head}
\label{sec:vis-head}
Every result above takes visibility from the front-end's forward--backward consistency test: follow
a point one frame forward and back again, and call it occluded if it fails to return to its start.
The tracker fills frames after a query with a forward sweep and frames before it with a backward
sweep, and each tests the round trip of its own hop: the forward sweep compares $\lVert
f_t+b_{t+1}\rVert$ against $\alpha(\lVert f_t\rVert+\lVert b_{t+1}\rVert)+\beta$, and the backward
sweep compares $\lVert f_{t-1}+b_t\rVert$ against the same expression in those two vectors. Write
$ok_t\in\{0,1\}$ for the outcome at frame $t$ --- $1$ when the residual is within tolerance --- and
$v_t\in\{0,1\}$ for the visibility the tracker actually reports, with $v_0=ok_0$.

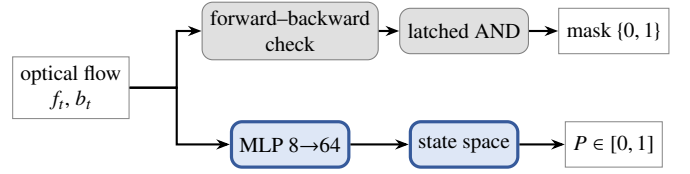
\begin{figure}[!tb]
\centering
\begin{tikzpicture}[font=\footnotesize,
  bx/.style={draw, minimum height=6mm, inner xsep=3pt, align=center},
  ob/.style={bx, sharp corners, fill=white, draw=black!45},
  pr/.style={bx, rounded corners, fill=black!12, draw=black!50},
  lr/.style={bx, rounded corners, fill=cTrain!18, draw=cTrain!75!black, very thick},
  a/.style={-{Stealth[length=1.8mm]}, thick}]
  \node[ob] (f)  at (0,0)       {optical flow\\$f_t$, $b_t$};
  \node[pr] (ck) at (2.9,0.75)   {forward--backward\\check};
  \node[pr] (an) at (5.2,0.75)   {latched AND};
  \node[ob] (mk) at (7.18,0.75)   {mask $\{0,1\}$};
  \node[lr] (ml) at (2.9,-0.75)  {MLP $8{\rightarrow}64$};
  \node[lr] (mm) at (5.2,-0.75)  {state space};
  \node[ob] (pb) at (7.18,-0.75)  {$P\in[0,1]$};
  \coordinate (x) at (1.4,0);
  \draw[a] (f.east) -- (x) |- (ck.west);
  \draw[a] (f.east) -- (x) |- (ml.west);
  \draw[a] (ck) -- (an);  \draw[a] (an) -- (mk);
  \draw[a] (ml) -- (mm);  \draw[a] (mm) -- (pb);
\end{tikzpicture}
\caption{\textbf{Visibility computation} (\S\ref{sec:vis-head}). Both paths read one pair of flow
  vectors per point per frame and nothing else --- no depth, no appearance, nothing from the
  refiner. \emph{Upper, the rule every earlier result was scored with}: follow the point one frame
  forward by $f_t$ and back by $b_{t+1}$, and call it occluded if it misses its start by more than
  $\alpha(\lVert f_t\rVert+\lVert b_{t+1}\rVert)+\beta$, with $\alpha$ and $\beta$ two hand-chosen
  constants ($0.05$ and $1$\,px) rather than anything fitted; the backward sweep applies the same
  test to $f_{t-1}$ and $b_t$. The result is latched, $v_{t+1}=v_t\wedge ok_t$, and one failure
  therefore fixes the point as occluded for every later frame. \emph{Lower, the learned head}: eight
  scalars per point per frame, all functions of $f_t$ and $b_t$, embedded to $64$; two state space
  layers mix along time only, never across points; a linear
  head and a sigmoid follow. The outputs differ in kind --- one bit against a probability, read as
  visible above $0.5$ --- and only the head can mark a point visible again after an occlusion, which
  $52$--$66\%$ of ground-truth points require.}
\label{fig:vis-head}
\end{figure}

Two properties of the rule limit it (Fig.~\ref{fig:vis-head}). Its tolerance is governed by two
hand-chosen numbers, $\alpha=0.05$ and $\beta=1$\,px: the same pair for every scene, every point and
every kind of motion, never fitted to data. The reported flag is also \emph{latched},
$v_{t+1}=v_t\wedge ok_t$, the logical AND of the two: a single failure fixes a point as occluded
for every later frame --- yet $52$--$66\%$ of ground-truth points on the three subsets reappear after
being occluded, and the rule recovers none of them.

We therefore learn the channel from the same evidence. Per point per frame the head reads two flow
vectors sampled at the tracked position: the forward flow $f_t$, which maps frame $t$ to $t{+}1$,
and the backward flow $b_t$, which maps $t$ to $t{-}1$. From these it builds eight numbers --- four
vector components and four lengths:
the two components of $f_t$ and the two of $b_t$;
the lengths $\lVert f_t\rVert$ and $\lVert b_t\rVert$;
and the two round-trip errors $\lVert f_t+b_{t+1}\rVert$, for the hop leaving frame $t$, and
$\lVert f_{t-1}+b_t\rVert$, for the hop arriving at it.
The last two are exactly the residuals the hand-written test thresholds, one per sweep direction,
handed over raw and without a tolerance. They carry nothing the two vectors do not already hold, and
are supplied directly because a sum of two separate inputs is awkward for a small network to form on
its own. An MLP embeds the eight to $64$, two state space layers mix along \emph{frames only} --- never
across points, each point being judged on its own track --- and a linear head with a sigmoid emits
$P(\text{visible})$, read as visible above $0.5$.
The head is $0.17$\,M parameters, sees no depth and no appearance, and reads nothing from the
refiner; it can therefore be trained standalone, and a change to the 3D refiner cannot invalidate it. Because
it decides each frame from its own evidence it has no latch, and a point that disappears behind
something can be marked visible again when it reappears.

\subsection{Training}
\label{sec:track-training}
Three modules are trained and nothing else: the 3D refiner, the depth scale refiner (with the
de-flicker stage it replaces) and the visibility head. The flow, depth and DINOv3 networks stay
frozen throughout.

The two depth-correcting stages share one recipe. We minimise an $\ell_1$ loss on 3D position over
visible points, normalised by the per-clip median anchor-frame depth so that the objective is
scale-balanced across the roughly $1$\,m (indoor) to $20$\,m (driving) range of the data. Training
uses AdamW~\cite{adamw} at learning rate $3\!\times\!10^{-4}$ for $20$k steps on $8$-frame windows in
bf16 ($16$-bit brain float).

The visibility head is trained separately, on its own objective and its own window. It reads nothing
from the other two (\S\ref{sec:vis-head}) and therefore needs neither their loss nor their $8$-frame
window: we minimise a binary cross-entropy against ground-truth visibility, weighted to offset the
class imbalance, for $3$k steps at learning rate $1\!\times\!10^{-3}$ over whole clips of up to $300$
frames. The long window is what lets the head see a point disappear and return inside one training
example, which an $8$-frame window rarely contains.

\section{Evaluation}
\label{sec:eval}

\begin{table*}[!tb]
\centering
\caption{TAPVid-3D minival, median-scaled 3D-AJ (standard leaderboard metric; higher better) ---
         \textbf{DA3-l depth} (\texttt{DA3Metric-Large}, 0.35\,B). Bold = best per column. In the row names, \emph{vmamba3} is the
           appearance- and geometry-conditioned refiner of \S\ref{sec:track-vmamba3-refiner} (the \emph{v} is for vision) and \emph{vmamba3-2pool} is that
           refiner with its temporal mixer replaced by VSSD-2pool (\ref{sec:vmamba3-noncausal}). All numbers
         are produced by our own evaluation pipeline on the same hardware. DA3-g counterpart:
         Table~\ref{tab:median-g}.}
\label{tab:median}
\small
\begin{tabular}{lrrrr}
\toprule
Method & drivetrack & pstudio & adt & mean \\
\midrule
\multicolumn{5}{l}{\emph{Published methods (external)}} \\
SpatialTrackerV2\,$^{*}$         & 0.017 & 0.012 & 0.025 & 0.018 \\
DELTA + DA3-l\,$^{\S}$             & \textbf{0.130} & \textbf{0.141} & 0.152 & \textbf{0.141} \\
DELTAv2 + DA3-l\,$^{\S}$           & \textbf{0.130} & 0.136 & 0.155 & 0.140 \\
TAPIP3D + DA3-l\,$^{\dagger}$      & 0.065 & 0.023 & 0.004 & 0.030 \\
TAPIP3D + MegaSaM\,$^{\ddagger}$ & 0.000 & 0.003 & 0.004 & 0.002 \\
TrackCraft3R\,$^{\P}$            & 0.003 & 0.013 & 0.010 & 0.009 \\
\midrule
\multicolumn{5}{l}{\emph{Ours (DA3-l depth)}} \\
SEA-RAFT + DA3-l (baseline)        & 0.108 & 0.111 & 0.132 & 0.117 \\
SEA-RAFT + MegaSaM               & 0.103 & 0.135 & \textbf{0.163} & 0.134 \\
SEA-RAFT+DA3-l+mamba3            & 0.057 & 0.041 & 0.123 & 0.074 \\
SEA-RAFT+MegaSaM+mamba3        & 0.059 & 0.048 & 0.119 & 0.075 \\
SEA-RAFT+DA3-l+vmamba3               & 0.093 & 0.057 & 0.141 & 0.097 \\
SEA-RAFT+DA3-l+vmamba3-2pool & 0.101 & 0.083 & 0.132 & 0.105 \\
WAFT+DA3-l            & 0.111 & 0.111 & 0.142 & 0.121 \\
WAFT+DA3-l+vmamba3  & 0.109 & 0.093 & 0.149 & 0.117 \\
WAFT+DA3-l+vmamba3-2pool & 0.110 & 0.096 & 0.157 & 0.121 \\
\quad $+$ visibility head & 0.090 & 0.082 & 0.133 & 0.102 \\
\bottomrule
\end{tabular}
\\[2pt]
{\footnotesize $^{*}$SpatialTrackerV2, $^{\S}$DELTA$+$DA3-l and its successor DELTAv2$+$DA3-l,
$^{\dagger}$TAPIP3D$+$DA3-l, $^{\ddagger}$TAPIP3D$+$MegaSaM, $^{\P}$TrackCraft3R are discussed in the
text; see \S\ref{sec:reliability}.}
\end{table*}

\begin{table*}[!tb]
\centering
\caption{TAPVid-3D minival, absolute metric-AJ (no scaling, fixed-meter thresholds 1\,cm--2.56\,m;
higher better) --- \textbf{DA3-l depth} (\texttt{DA3Metric-Large}, 0.35\,B). Bold = best per column.
Row names and evaluation conditions as in Table~\ref{tab:median}.
\textbf{Highest mean across all four tables: WAFT+DA3-l+vmamba3-2pool with the visibility head}
($0.256$); no DA3-g pipeline reaches it (\S\ref{sec:da3lg}). DA3-g counterpart: Table~\ref{tab:abs-g}.}
\label{tab:abs}
\small
\begin{tabular}{lrrrr}
\toprule
Method & drivetrack & pstudio & adt & mean \\
\midrule
\multicolumn{5}{l}{\emph{Published methods (external)}} \\
SpatialTrackerV2\,$^{*}$         & 0.008 & 0.192 & 0.179 & 0.126 \\
DELTA + DA3-l\,$^{\S}$             & 0.006 & 0.199 & 0.344 & 0.183 \\
DELTAv2 + DA3-l\,$^{\S}$           & 0.005 & 0.190 & \textbf{0.345} & 0.180 \\
TAPIP3D + DA3-l                    & 0.006 & 0.131 & 0.164 & 0.100 \\
TAPIP3D + MegaSaM\,$^{\dagger}$  & 0.000 & 0.000 & 0.000 & 0.000 \\
TrackCraft3R\,$^{\P}$            & 0.002 & 0.025 & 0.032 & 0.020 \\
\midrule
\multicolumn{5}{l}{\emph{Ours (DA3-l depth)}} \\
SEA-RAFT + DA3-l (baseline)        & 0.005 & 0.165 & 0.273 & 0.147 \\
SEA-RAFT + MegaSaM               & 0.000 & 0.082 & 0.209 & 0.097 \\
SEA-RAFT+DA3-l+mamba3            & 0.082 & 0.186 & 0.272 & 0.180 \\
SEA-RAFT+MegaSaM+mamba3        & 0.001 & 0.092 & 0.208 & 0.100 \\
SEA-RAFT+DA3-l+vmamba3               & 0.130 & 0.248 & 0.298 & 0.225 \\
SEA-RAFT+DA3-l+vmamba3-2pool & 0.135 & 0.271 & 0.299 & 0.235 \\
WAFT+DA3-l            & 0.005 & 0.166 & 0.299 & 0.157 \\
WAFT+DA3-l+vmamba3  & \textbf{0.141} & 0.276 & 0.328 & 0.248 \\
WAFT+DA3-l+vmamba3-2pool & \textbf{0.141} & 0.285 & 0.338 & 0.255 \\
\quad $+$ visibility head & 0.132 & \textbf{0.314} & 0.323 & \textbf{0.256} \\
\bottomrule
\end{tabular}
\\[2pt]
{\footnotesize
$^{*}$SpatialTrackerV2 evaluated with $s_{\text{wind}}{=}60$ on RTX\,4080\,Laptop (12\,GB).
$^{\S}$DELTA$+$DA3-l and DELTAv2$+$DA3-l hold the best \emph{adt} score of any method between them
($0.345$, $0.344$), but their far-field \emph{drivetrack} depth scale keeps their mean below ours.
$^{\dagger}$TAPIP3D$+$MegaSaM scores exactly $0$ everywhere: MegaSaM's bundle adjustment is degenerate
without parallax, and its scale therefore drifts far from the monocular prior (a $17\times$ error on
drivetrack) and the 3D error exceeds even the $256$\,cm threshold --- an evaluation-condition mismatch, not a failure of
TAPIP3D. $^{\P}$TrackCraft3R on all 150 clips; see Table~\ref{tab:median}.}
\end{table*}

\begin{table*}[!tb]
\centering
\caption{TAPVid-3D minival, median-scaled 3D-AJ (leaderboard metric; higher better) ---
         \textbf{DA3-g depth} (\texttt{DA3Nested-Giant-Large}, 1.40\,B): every DA3-consuming method
         re-run on the DA3-g backbone, all other components unchanged. Bold = best per column.
         ``OOM'' = TAPIP3D adt out-of-memory. DA3-l counterpart: Table~\ref{tab:median}.}
\label{tab:median-g}
\small
\begin{tabular}{lrrrr}
\toprule
Method (DA3-g) & drivetrack & pstudio & adt & mean \\
\midrule
\multicolumn{5}{l}{\emph{Published methods (external)}} \\
DELTA + DA3-g            & \textbf{0.137} & \textbf{0.049} & 0.141 & \textbf{0.109} \\
DELTAv2 + DA3-g          & 0.131 & 0.046 & \textbf{0.148} & 0.108 \\
SpatialTrackerV2 + DA3-g & 0.018 & 0.008 & 0.027 & 0.017 \\
TAPIP3D + DA3-g          & 0.057 & 0.011 & OOM   & 0.034 \\
\midrule
\multicolumn{5}{l}{\emph{Ours}} \\
SEA-RAFT + DA3-g (baseline)              & 0.112 & 0.036 & 0.120 & 0.089 \\
WAFT + DA3-g (no refiner)                & 0.116 & 0.036 & 0.130 & 0.094 \\
WAFT+DA3-g+vmamba3-2pool                 & 0.116 & 0.032 & 0.129 & 0.093 \\
\quad $+$ per-frame scale head            & 0.115 & 0.035 & 0.126 & 0.092 \\
WAFT+DA3-g+deflicker (standalone)        & 0.116 & 0.036 & 0.130 & 0.094 \\
WAFT+DA3-g+deflicker+vmamba3-2pool & 0.115 & 0.034 & 0.126 & 0.092 \\
WAFT+DA3-g+scale refiner+2pool & 0.111 & 0.033 & 0.128 & 0.091 \\
\quad $+$ visibility head & 0.091 & 0.031 & 0.108 & 0.076 \\
\bottomrule
\end{tabular}
\end{table*}

\begin{table*}[!tb]
\centering
\caption{TAPVid-3D minival, absolute metric-AJ (fixed-meter thresholds; higher better) ---
         \textbf{DA3-g depth} (\texttt{DA3Nested-Giant-Large}, 1.40\,B): every DA3-consuming method
         re-run on the DA3-g backbone, all other components unchanged. Bold = best per column.
         \textbf{Best DA3-g mean: our WAFT+DA3-g+scale refiner+2pool with the visibility head
         ($0.229$)}, which \S\ref{sec:results} reports as matching, not exceeding, the best external
         DA3-g method, DELTA$+$DA3-g ($0.227$). Every DA3-g configuration here remains below the
         DA3-l best ($0.256$, Table~\ref{tab:abs}), which is this table's purpose.
         ``OOM'' = TAPIP3D adt out-of-memory. DA3-l counterpart: Table~\ref{tab:abs}.}
\label{tab:abs-g}
\small
\begin{tabular}{lrrrr}
\toprule
Method (DA3-g) & drivetrack & pstudio & adt & mean \\
\midrule
\multicolumn{5}{l}{\emph{Published methods (external)}} \\
DELTA + DA3-g            & \textbf{0.164} & 0.194 & \textbf{0.324} & 0.227 \\
DELTAv2 + DA3-g          & 0.159 & 0.171 & 0.319 & 0.216 \\
SpatialTrackerV2 + DA3-g & 0.043 & 0.184 & 0.170 & 0.132 \\
TAPIP3D + DA3-g          & 0.110 & 0.148 & OOM   & 0.129 \\
\midrule
\multicolumn{5}{l}{\emph{Ours}} \\
SEA-RAFT + DA3-g (baseline)              & 0.136 & 0.175 & 0.269 & 0.193 \\
WAFT + DA3-g (no refiner)                & 0.144 & 0.177 & 0.295 & 0.205 \\
WAFT+DA3-g+vmamba3-2pool                 & 0.152 & 0.207 & 0.307 & 0.222 \\
\quad $+$ per-frame scale head            & 0.153 & 0.204 & 0.307 & 0.221 \\
WAFT+DA3-g+deflicker (standalone)        & 0.119 & 0.181 & 0.300 & 0.200 \\
WAFT+DA3-g+deflicker+vmamba3-2pool & 0.163 & 0.203 & 0.306 & 0.224 \\
WAFT+DA3-g+scale refiner+2pool & 0.163 & 0.209 & 0.310 & 0.227 \\
\quad $+$ visibility head & 0.154 & \textbf{0.234} & 0.299 & \textbf{0.229} \\
\bottomrule
\end{tabular}
\end{table*}

\subsection{Evaluation dataset}
\label{sec:dataset}
We evaluate on the TAPVid-3D~\cite{tapvid3d} \emph{minival} split ($150$ clips: $50$ each from
DriveTrack~\cite{drivetrack}, Panoptic Studio~\cite{panopticstudio}, and Aria Digital
Twin~\cite{ariadigitaltwin}). Both splits we use are the benchmark authors' own published file lists,
not partitions of our choosing: minival and the larger \texttt{full\_eval} ($4419$ clips) are disjoint by
construction, and we train the refiner on \texttt{full\_eval} and test on minival; no clip appears in
both roles. We refer to
the three subsets by their lower-case informal names throughout: \emph{drivetrack}, \emph{pstudio}, and
\emph{adt}. TAPVid-3D is the standard benchmark for metric 3D point tracking, and its three subsets span a
range of conditions and scales relevant to real applications: drivetrack is outdoor driving ($10$--$50$\,m),
pstudio is indoor multi-person motion-capture ($1$--$5$\,m), and adt is egocentric indoor motion
($\sim$$1$\,m).

\subsection{Evaluation protocol}
\label{sec:protocol}
All methods are run by us on a single RTX~4080 Laptop GPU ($12$\,GB) so that hardware is held constant;
training the refiner takes about $9.5$\,h on this GPU. We report two families of metrics. The \emph{normalized} 3D-AJ,
TAPVid-3D's standard tracking metric, is the standard leaderboard metric, and it is
forgiving of metric scale in two complementary ways. First, it rescales each prediction by a single global
factor---the median ratio between predicted and ground-truth depth. This is what removes the overall
scale: it cancels any constant, scene-wide error; only the relative shape of the track is scored, not
its size. Second, the correctness thresholds grow with each point's depth rather than being fixed
distances in meters, and a far point is therefore allowed a proportionally larger error than a near one. This
forgives errors that \emph{vary} with depth, which the single global factor cannot correct (for instance
depths that drift increasingly wrong with distance). Together the two make the metric reward relative
structure while saying nothing about whether the recovered depths are metrically correct. The
\emph{absolute} metric-AJ removes the median rescaling and scores at fixed metric thresholds
$\{1,4,16,64,256\}$\,cm; it therefore reflects real-world 3D error in meters---the quantity a downstream metric
application actually needs, and the reason we report it alongside the standard leaderboard metric. We
validate that our scorer is faithful: it reproduces SpatialTracker's~\cite{spatialtracker} published
DriveTrack 3D-AJ to three decimals once the intrinsics convention is matched.

We report results separately for two frozen Depth-Anything-3~\cite{da3} metric-depth backbones that differ
by $4\times$ in parameter count: \emph{DA3-l} (\texttt{DA3Metric-Large}, $0.35$\,B, used for all our
primary results) and \emph{DA3-g} (\texttt{DA3Nested-Giant-Large}, $1.40$\,B). We split the tables and
figures for this reason alone: mixing two pipelines that differ $4\times$ in depth-backbone size into one
ranking would compare unequal parameter budgets as if they were interchangeable. We therefore give each
backbone its own pair of tables and figures (normalized then absolute) in \S\ref{sec:results}; every
ranking stays parameter-matched. The two backbones also fail in
different, complementary ways, which is why the larger one does not simply dominate the smaller.

\subsection{Evaluation results}
\label{sec:results}
\subsubsection{DA3-l depth (0.35\,B): leaderboard results}
\label{sec:res-da3l-norm}
Table~\ref{tab:median} reports the normalized metric. Here a strong depth baseline
(SEA-RAFT\,+\,MegaSaM) and the feed-forward trackers DELTA and its successor DELTAv2 lead; our learned
variants rank in the middle of the table on this metric. We state this plainly: \emph{we do not claim
state of the art on the scale-invariant leaderboard.}

\subsubsection{DA3-l depth: absolute-metric results}
\label{sec:res-da3l-abs}
Table~\ref{tab:abs} reports the absolute metric that a downstream 3D application
actually needs. The ranking reverses relative to the leaderboard metric (Table~\ref{tab:median}): our best configuration
(WAFT flow\,+\,DA3-l depth\,+\,the vmamba3 refiner) attains the highest mean metric-AJ,
$0.256$ with the VSSD-2pool mixer and the flow-only visibility head of \S\ref{sec:vis-head},
$0.255$ with that mixer and the front-end's forward--backward rule, and $0.248$ with one pool, and
is best on two of three subsets, ahead of every external method evaluated under the same
conditions. Relative to
the DA3-l baseline the refiner lifts mean metric-AJ from $0.147$ to $0.256$ and roughly halves mean 3D error;
relative to the best feed-forward external (DELTA, mean $0.183$; its successor DELTAv2 close behind at
$0.180$) it is ahead overall, the DELTA family retaining only the Aria subset.

\subsubsection{DA3-g depth (1.40\,B): leaderboard results}
\label{sec:res-da3g-norm}
Table~\ref{tab:abs} shows every DA3-l-based method---including ours---collapsing on far-field drivetrack
(metric-AJ $\le0.141$); \S\ref{sec:da3lg} explains why. We therefore re-run every DA3-l-based
method---external and ours---on the DA3-g backbone, re-caching depth and re-training our refiner from
scratch on it, everything else unchanged. Table~\ref{tab:median-g} reports the
normalized metric on DA3-g. The ranking is essentially unchanged from DA3-l (Table~\ref{tab:median}):
DELTA, DELTAv2, and the training-free baselines still lead this metric.

\subsubsection{DA3-g depth: absolute-metric results}
\label{sec:res-da3g-abs}
Table~\ref{tab:abs-g} reports the absolute metric on DA3-g. Drivetrack
rises across the board relative to DA3-l (e.g.\ our refiner $0.141\to0.163$). Pstudio and adt,
however, \emph{fall} instead (our refiner $0.285\to0.203$ and $0.338\to0.306$), leaving every DA3-g
pipeline below the DA3-l one.

Two stages close the gap to the strongest external DA3-g method. Reading the depth map directly
rather than through the tracker (the scale refiner of \S\ref{sec:scale-refiner}) takes the mean from
$0.224$ to $0.227$, and replacing the forward--backward visibility rule with the flow-only head of
\S\ref{sec:vis-head} takes it to $0.229$, against DELTA's $0.227$ (DELTAv2 behind at $0.216$).
\emph{We report this as matching DELTA rather than exceeding it}: the margin is $0.0013$, and two
seeds of one of our own arms differ by $0.0056$; a gap this far inside a single arm's run-to-run
spread is not an ordering. The same substitution is what produces our best absolute figure on DA3-l,
$0.256$ (Table~\ref{tab:abs}).

The visibility head is a genuine trade rather than an improvement at no cost, and the normalised tables show the
other side of it: on DA3-g it moves the leaderboard metric from $0.091$ to $0.076$ and on DA3-l from
$0.121$ to $0.102$ (Tables~\ref{tab:median-g}, \ref{tab:median}), losing on every subset of both
backbones. Marking more points visible exposes more of them to the position test, and the normalised
metric median-scales before thresholding, which makes its thresholds relative and much tighter. A reader
optimising the leaderboard metric should keep the forward--backward rule; the absolute figures we
report use the head.

\subsection{Ablation}
\label{sec:ablation}
Tables~\ref{tab:median}--\ref{tab:abs} can also be read as an ablation on DA3-l depth, because their rows
add one component at a time. Adding the plain Mamba-3 depth
refiner (SEA-RAFT+DA3-l+mamba3) to the DA3-l baseline already improves absolute accuracy; conditioning it
on DINOv3 and the local depth patch (the vmamba3 refiner, SEA-RAFT+DA3-l+vmamba3) adds a further
large gain (mean metric-AJ $0.147\!\to\!0.225$); swapping the SEA-RAFT front-end for WAFT
(WAFT+DA3-l+vmamba3) lifts every subset again to $0.248$, and a second independently-read pool
(VSSD-2pool, \ref{sec:vmamba3-noncausal}) adds $0.007$ more, reaching $0.255$ for
$38\%$ more refiner parameters; the flow-only visibility head of \S\ref{sec:vis-head} adds the final
$0.001$, giving the $0.256$ reported above.

Tables~\ref{tab:median-g}--\ref{tab:abs-g} add the same components in the same order on DA3-g
depth: the vmamba3 refiner outperforms the no-refiner baseline (mean absolute metric-AJ
$0.205\!\to\!0.222$), and the two-stage de-flicker-then-refine pipeline (\S\ref{sec:scale-refiner})
adds a further increment, to $0.224$. The per-frame scale head does not: at two
pools it moves the mean by $-0.001$; the de-flicker stage rather than the scale head is what
addresses DA3-g's per-frame scale drift. Reading the depth map directly instead of through the
tracker (the scale refiner of \S\ref{sec:scale-refiner}) reaches $0.227$, and the flow-only
visibility head takes the best DA3-g mean to $0.229$, which \S\ref{sec:results} reports as matching
DELTA$+$DA3-g ($0.227$) rather than exceeding it. Every DA3-g configuration remains below the DA3-l
ceiling.

Every number in this paper is a single training run, and we report no seed variance for any of them:
these ladders are not repeated, and nothing here therefore bounds what a rerun would move. The smaller
steps---the $0.006$ the second pool adds, and the $-0.001$ the scale head moves---should therefore be read
as measurements rather than as demonstrated orderings. We treat the larger steps, such as the
$0.147\!\to\!0.225$ the appearance-conditioned refiner produces, as real rather than noise, because a
gap of that size would be unusual for this kind of run-to-run variation; that is a judgement based on
experience with similar training runs, not a claim established by repeats we have not run.

\subsection{Efficiency}
\label{sec:eval-efficiency}
Table~\ref{tab:efficiency} reports throughput and trainable parameters on the DA3-l backbone. The refiner
adds under $1$\,M parameters and keeps the pipeline at $6$--$12$\,fps: comparable in throughput to the
DELTA family, $1.6$--$2\times$ faster than TAPIP3D and SpatialTrackerV2, and two orders of magnitude
faster than the video-diffusion tracker TrackCraft3R. \S\ref{sec:discussion} explains why the external methods score as low as
Table~\ref{tab:median} shows under this hardware budget.

\begin{table}[!tb]
\centering
\caption{Inference throughput on RTX\,4080 (12\,GB), measured on TAPVid-3D minival with the
         \textbf{DA3-l} depth backbone (\texttt{DA3Metric-Large}, 0.35\,B). Our lightweight SSM
         refiner adds modest overhead over the baseline while remaining comparable in throughput to
         the DELTA family, $1.6$--$2\times$ faster than TAPIP3D and SpatialTrackerV2, and two orders of
         magnitude faster than the video-diffusion tracker TrackCraft3R; its learned module ($0.44$\,M, $0.61$\,M with the
         two-pool mixer, $0.79$\,M with the visibility head) is $40$--$3000\times$ smaller than the external
         trackers ($25.8$\,M--$1.3$\,B). External param counts are summed from the official released
         checkpoints (no paper states them). Bold marks the best value in each column. Throughput is an end-to-end average over the evaluation run; a row whose
         figure is instead carried from another row, measured on one subset, or derived is
         marked in the note below.
         DA3-g counterpart: Table~\ref{tab:efficiency-g}.}
\label{tab:efficiency}
\footnotesize\setlength{\tabcolsep}{4pt}
\begin{tabular}{lrr}
\toprule
Method & fps & trainable params \\
\midrule
\multicolumn{3}{l}{\emph{Published methods (external)}} \\
TAPIP3D$^{\star}$                      & 3.1  & $25.8$\,M \\
SpatialTrackerV2 (s\_wind\,=\,60)      & 4.0  & ${\approx}1.23$\,B$^{\ddagger}$ \\
DELTA (DenseTrack3D)$^{\star}$         & 7.7  & $59.2$\,M \\
DELTAv2 (DenseTrack3Dv2)$^{\star}$     & 6.9  & $51.4$\,M \\
TrackCraft3R$^{\S}$                    & 0.06 & $1.3$\,B$^{\S}$ \\
\midrule
\multicolumn{3}{l}{\emph{Ours \& training-free baselines}} \\
DA3-l depth (shared precompute)$^{\star}$ & 10   & \textbf{0} \\
SEA-RAFT $+$ DA3-l (baseline)$^{\star}$   & 12.4            & \textbf{0} \\
WAFT $+$ DA3-l$^{\star}$                  & 7.7  & \textbf{0} \\
SEA-RAFT+DA3-l+mamba3$^{\star}$           & \textbf{12.5}   & $0.40$\,M \\
SEA-RAFT+DA3-l+vmamba3$^{\star,\dagger}$  & 7.7             & $0.44$\,M \\
WAFT+DA3-l+vmamba3$^{\star}$             & 6.3  & $0.44$\,M \\
SEA-RAFT+DA3-l+vmamba3-2pool$^{\star,\dagger,\P}$ & 7.7           & $0.61$\,M \\
WAFT+DA3-l+vmamba3-2pool$^{\star,\P}$    & 6.3  & $0.61$\,M \\
\quad $+$ visibility head$^{\star,\P,\natural}$ & 6.3  & $0.79$\,M \\
\bottomrule
\end{tabular}
\\[2pt]
{\footnotesize
$^{\star}$Precomputed DA3-l depth from cache (not counted in fps); the preprocessing row is a
one-time offline step shared by all such methods.
$^{\dagger}$SEA-RAFT+DA3-l+vmamba3 throughput measured on the drivetrack subset only.
$^{\ddagger}$SpatialTrackerV2 trains end-to-end: a VGGT-style front-end~\cite{vggt} ($1.16$\,B, initialised
from VGGT) plus a $69$\,M Track3D head.
$^{\P}$The two-pool rows carry their one-pool counterpart's throughput: the second pool adds
$0.17$\,M refiner parameters, no frozen network, and $1$--$3\%$ of throughput, within run-to-run
variation.
$^{\S}$TrackCraft3R is a pre-trained $1.3$\,B video-diffusion model (Wan2.1 DiT), LoRA
fine-tuned; the $1.3$\,B is the frozen backbone; throughput via windowed inference (see text).
$^{\natural}$The visibility head adds $0.17$\,M parameters beside a frozen $0.35$\,B depth backbone
and does not change throughput measurably; the row carries the preceding row's fps.}
\end{table}

Table~\ref{tab:efficiency-g} repeats this comparison on the DA3-g backbone: inference throughput is unchanged,
since depth is read from an offline cache either way, and the 3D refiner still trains under $1$\,M
parameters, but every method's \emph{total} footprint rises by DA3-g's $1.40$\,B; the comparison therefore stays
parameter-matched against the largest external trackers rather than obtained with an outsized model.

\begin{table}[!tb]
\centering
\caption{Inference throughput and model size on the \textbf{DA3-g} pipeline
         (\texttt{DA3Nested-Giant-Large}, 1.40\,B) --- identical to Table~\ref{tab:efficiency} except
         for the depth backbone. Online fps is unchanged from the DA3-l table because depth is read
         from the offline cache; only the one-time depth precompute is slower for the giant backbone.
Every method here reads the same $1.40$\,B DA3-g depth, so
         that backbone is common to all of them and is not what separates them; what differs is
         the trainable count, where ours ranges from $0.44$\,M for the refiner alone to $1.37$\,M
         with the scale refiner and visibility head added. Bold marks the best value in each column. Throughput is an end-to-end average over the evaluation run; a row whose
         figure is instead carried from another row, measured on one subset, or derived is
         marked in the note below.
         Columns and external counts as in Table~\ref{tab:efficiency}; TrackCraft3R could not be
         re-run on DA3-g ($12$\,GB OOM) and is omitted here.}
\label{tab:efficiency-g}
\footnotesize\setlength{\tabcolsep}{4pt}
\begin{tabular}{lrr}
\toprule
Method & fps & trainable \\
\midrule
\multicolumn{3}{l}{\emph{Published methods (external)}} \\
TAPIP3D + DA3-g$^{\star}$              & 3.1  & $25.8$\,M \\
SpatialTrackerV2 (s\_wind\,=\,60)      & 4.0  & ${\approx}1.23$\,B$^{\ddagger}$ \\
DELTA + DA3-g$^{\star}$                & \textbf{7.7}  & $59.2$\,M \\
DELTAv2 + DA3-g$^{\star}$              & 6.9  & $51.4$\,M \\
\midrule
\multicolumn{3}{l}{\emph{Ours \& training-free baselines}} \\
DA3-g depth (shared precompute)$^{\dagger}$ & 2.5 & \textbf{0} \\
WAFT $+$ DA3-g (no refiner)$^{\star}$       & \textbf{7.7} & \textbf{0} \\
WAFT+DA3-g+vmamba3$^{\star}$                & 6.3 & $0.44$\,M \\
WAFT+DA3-g+scale refiner+2pool$^{\star,\natural}$ & 6.3 & $1.20$\,M \\
\quad $+$ visibility head$^{\star,\natural}$      & 6.3 & $1.37$\,M \\
\bottomrule
\end{tabular}
\\[2pt]
{\footnotesize
$^{\star}$Depth read from the shared offline cache; not counted in fps.
$^{\dagger}$The DA3-g depth precompute is offline/one-time; the giant backbone has ${\approx}4\times$ the
parameters of DA3-l and precomputes correspondingly slower; its figure is derived from the
parameter ratio, not measured.
$^{\ddagger}$SpatialTrackerV2 trains end-to-end (VGGT-style $1.16$\,B front-end $+$ $69$\,M head; see
Table~\ref{tab:efficiency}); its own depth means it adds no DA3.
TAPIP3D and DELTA consume the \emph{same} shared DA3-g depth as our method; the comparison is therefore parameter-matched on the backbone and differs only in what each method trains on top of it.
$^{\natural}$The scale refiner and visibility head add $0.58$\,M and $0.17$\,M parameters beside a
frozen $1.4$\,B depth backbone and do not change throughput measurably; these rows carry the
preceding row's fps.}
\end{table}

\subsection{Qualitative results}
\label{sec:qual}
\begin{figure*}[p]
  \centering
  \makebox[\linewidth][c]{%
  \begin{subfigure}[b]{0.36\linewidth}\centering\includegraphics[width=\linewidth]{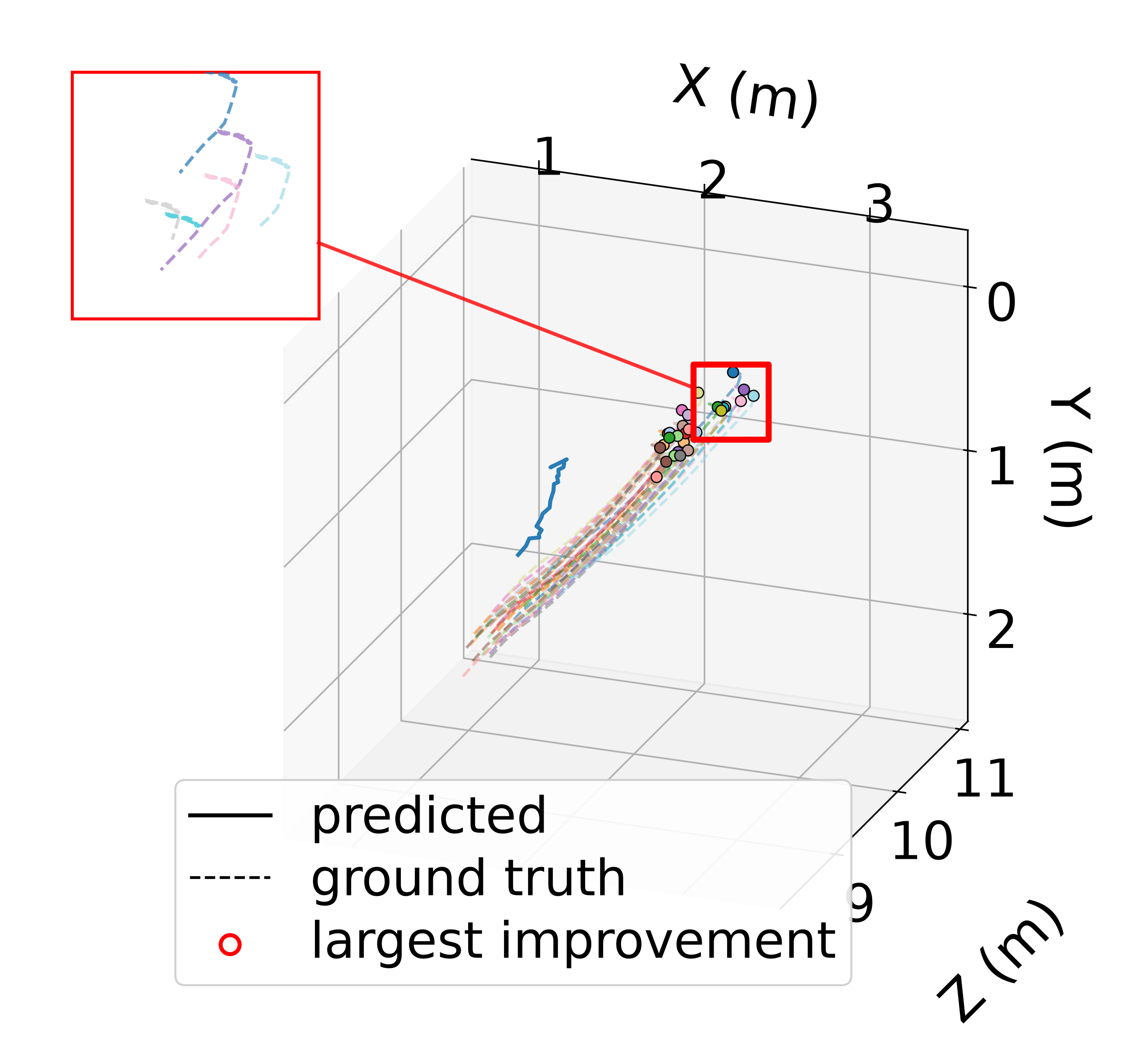}\\[1pt]{\footnotesize DELTA+DA3-l}\end{subfigure}%
  \hspace{3mm}%
  \begin{subfigure}[b]{0.36\linewidth}\centering\includegraphics[width=\linewidth]{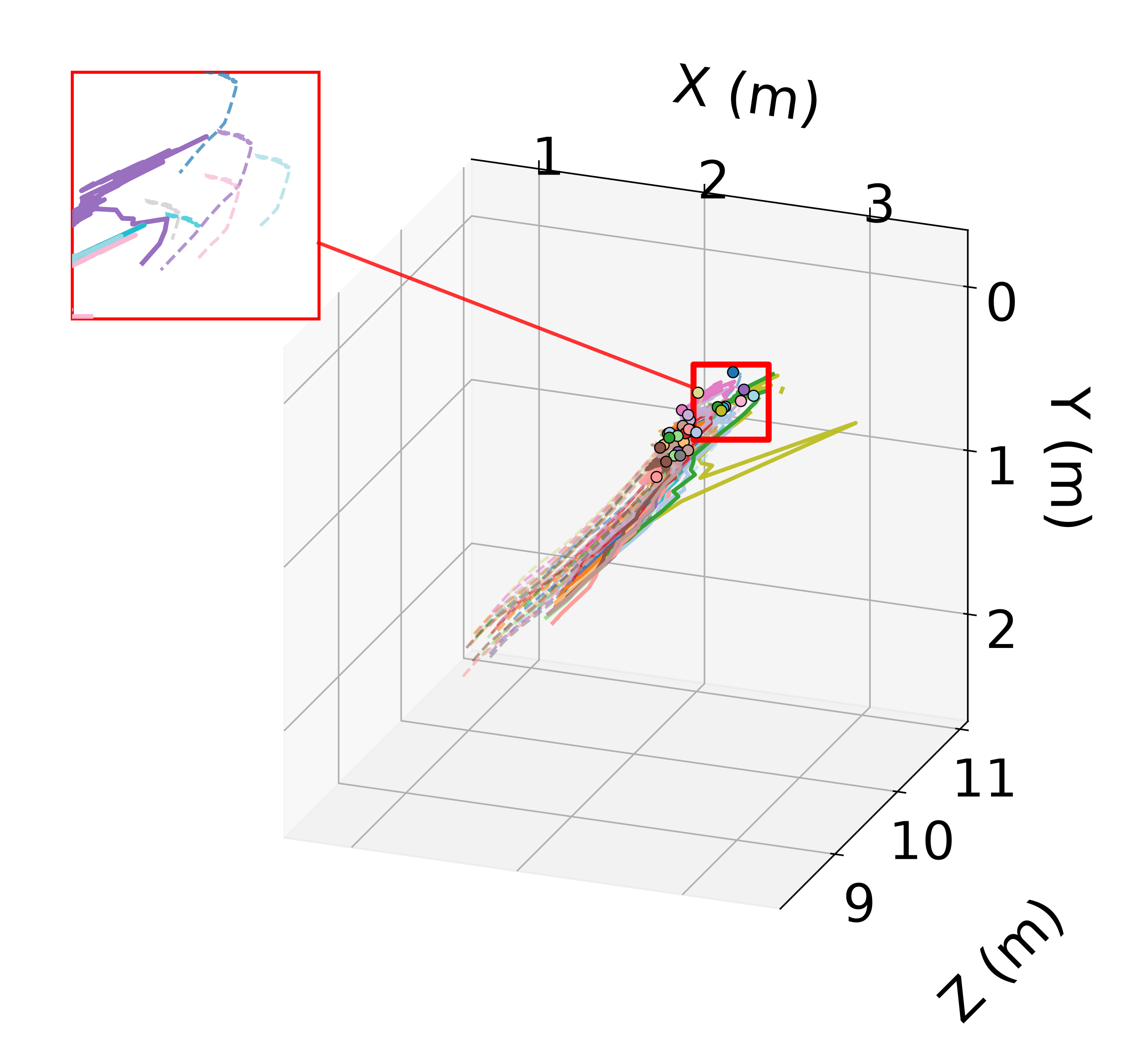}\\[1pt]{\footnotesize Ours (WAFT+DA3-l+vmamba3-2pool $+$ vis.\ head)}\end{subfigure}%
  }\\[1pt]
  {\footnotesize (a) drivetrack}\\[2pt]
  \makebox[\linewidth][c]{%
  \begin{subfigure}[b]{0.36\linewidth}\centering\includegraphics[width=\linewidth]{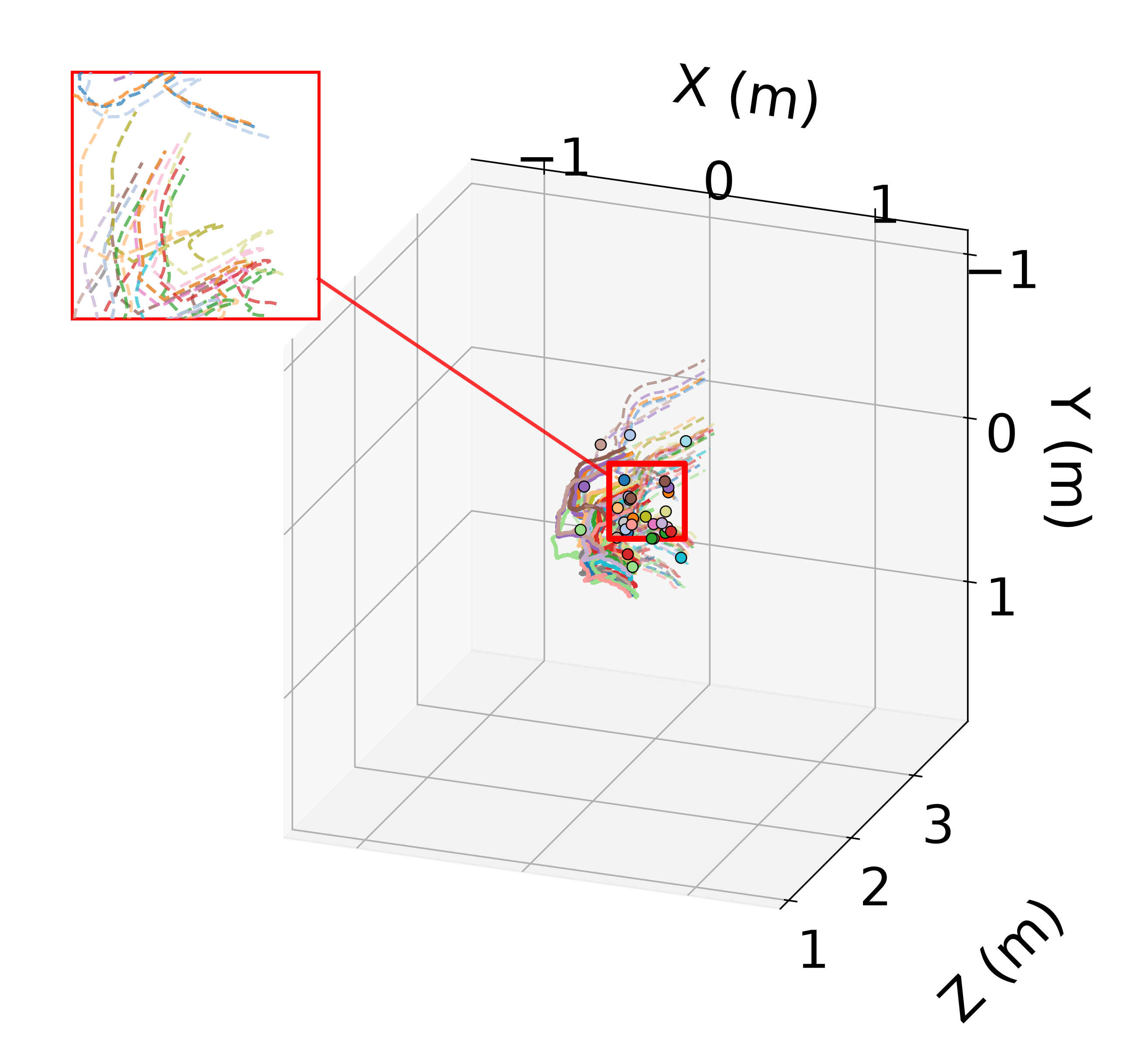}\\[1pt]{\footnotesize DELTA+DA3-l}\end{subfigure}%
  \hspace{3mm}%
  \begin{subfigure}[b]{0.36\linewidth}\centering\includegraphics[width=\linewidth]{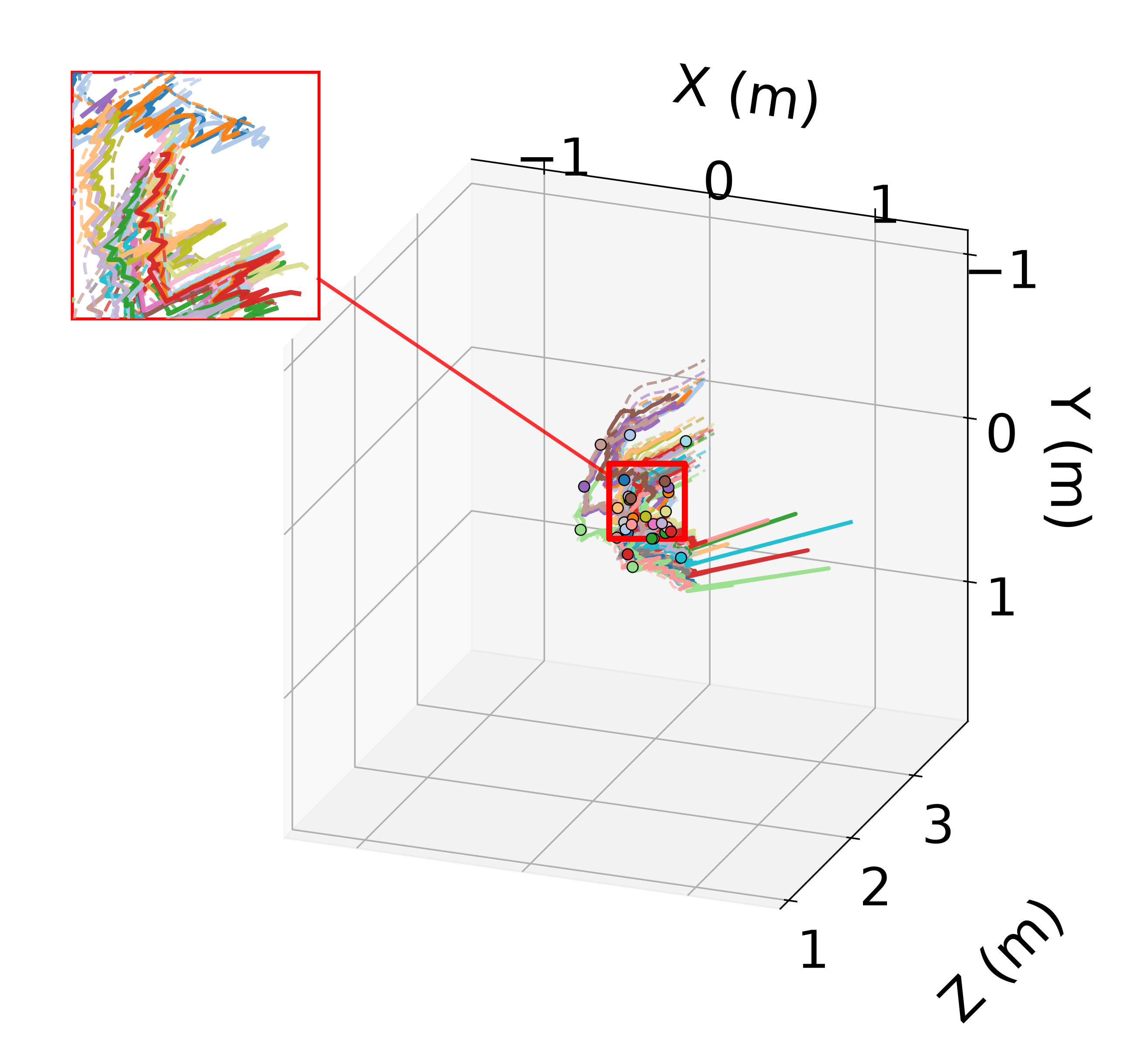}\\[1pt]{\footnotesize Ours (WAFT+DA3-l+vmamba3-2pool $+$ vis.\ head)}\end{subfigure}%
  }\\[1pt]
  {\footnotesize (b) pstudio}\\[2pt]
  \makebox[\linewidth][c]{%
  \begin{subfigure}[b]{0.36\linewidth}\centering\includegraphics[width=\linewidth]{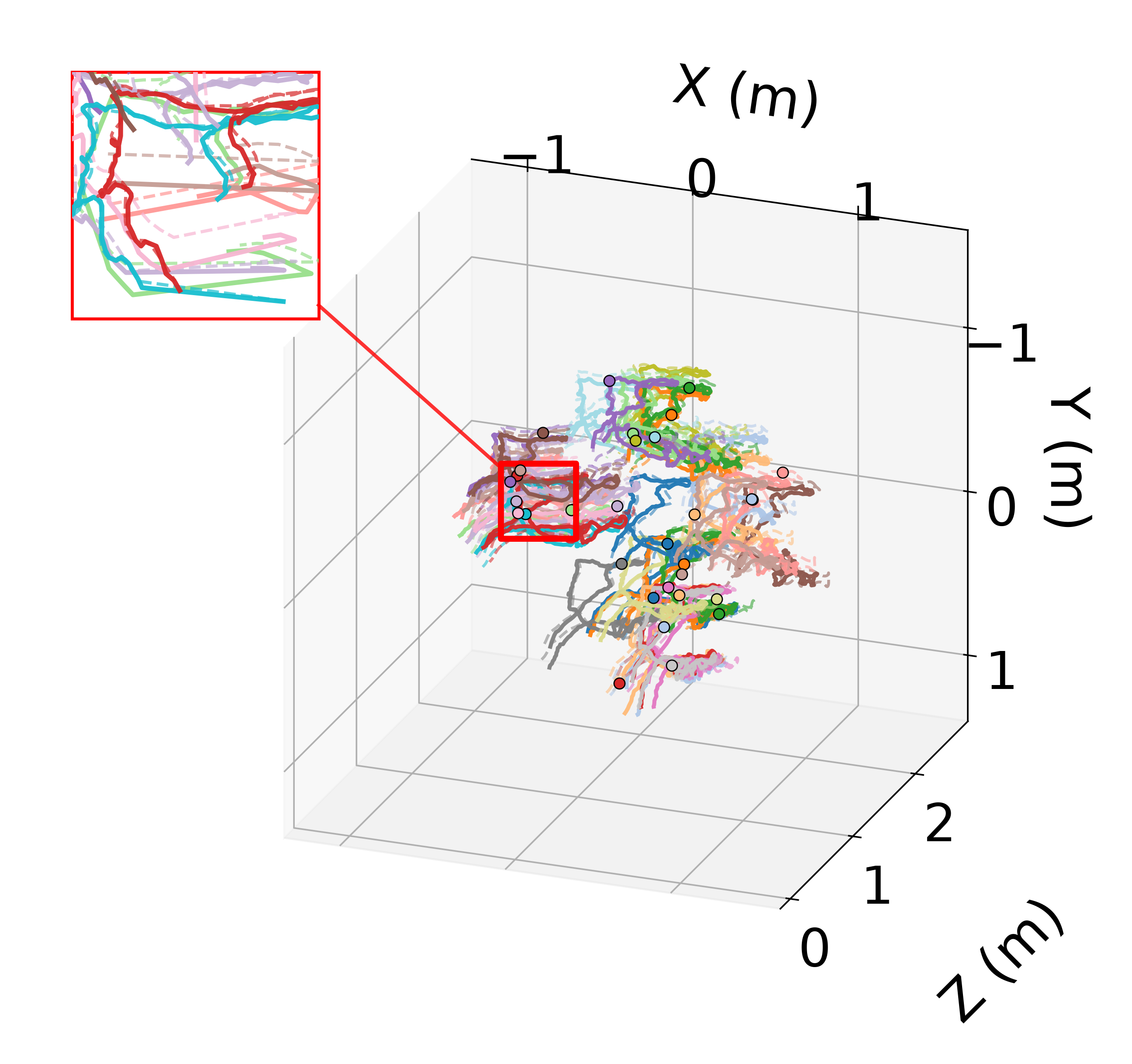}\\[1pt]{\footnotesize DELTA+DA3-l}\end{subfigure}%
  \hspace{3mm}%
  \begin{subfigure}[b]{0.36\linewidth}\centering\includegraphics[width=\linewidth]{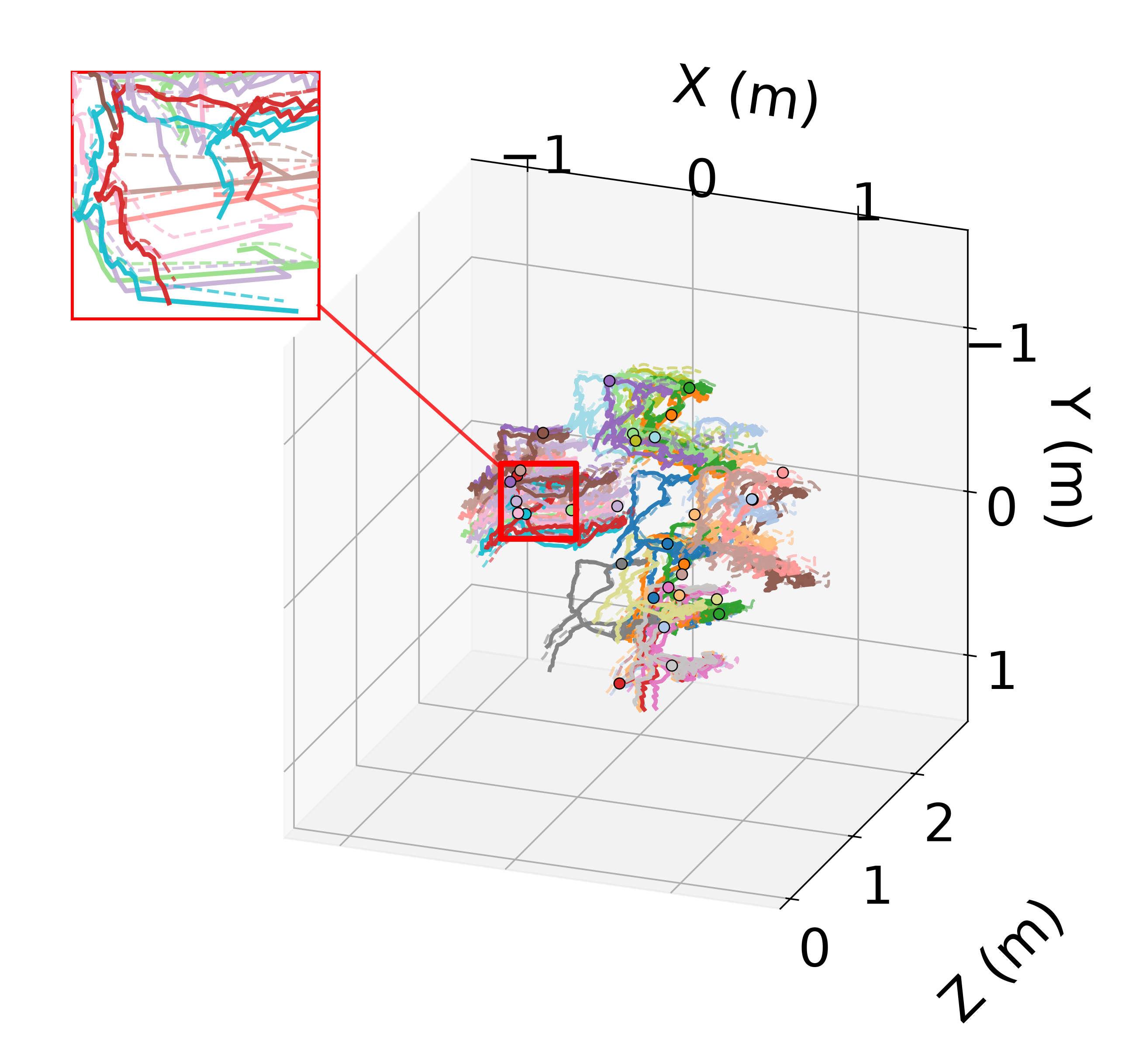}\\[1pt]{\footnotesize Ours (WAFT+DA3-l+vmamba3-2pool $+$ vis.\ head)}\end{subfigure}%
  }\\[1pt]
  {\footnotesize (c) adt}
  \caption{3D trajectory comparison, DELTA$+$DA3-l (left of each pair) against ours (right;
  the panel labels name the configuration), one clip per subset: (a) drivetrack, (b) pstudio,
  (c) adt. Both use the same DA3-l depth, which isolates the tracker. Solid = predicted,
  dashed = ground truth; a prediction that leaves the plotted volume simply does not appear. The red
  square marks the track our method improves the most over DELTA, the same one in both panels, and the
  inset shows that region at twice the size. The difference is plain in (a) and (b) but hard to see in
  (c): adt is the one subset of the three where DELTA scores above us (Table~\ref{tab:abs}).
  \S\ref{sec:qual} discusses each subset.}
  \label{fig:qual}
\end{figure*}

Figure~\ref{fig:qual} compares our best configuration --- WAFT flow $+$ DA3-l $+$ the VSSD-2pool
vmamba3 refiner with the flow-only visibility head --- against
the closest similarly-sized external method, DELTA$+$DA3-l, on one clip from each subset. On
\emph{drivetrack}, DELTA's far-field predictions collapse toward the camera and leave the scene volume
entirely---no solid track is visible in its panel---while ours stays on the ground truth; this is the
single largest source of our absolute-metric-AJ advantage on this subset. On \emph{pstudio}, our depth recovers
the mid-range metric depth DELTA misses. On \emph{adt}, DA3-l depth is already accurate at this near
range; both methods therefore do well, and the two panels are hard to tell apart. This is the one
subset of the three where DELTA is ahead of us (metric-AJ $0.344$ against our $0.323$,
Table~\ref{tab:abs}); our margins on drivetrack and pstudio are what carry the mean.

\section{Discussion}
\label{sec:discussion}

\subsection{DA3-l vs.\ DA3-g: far-field bias vs.\ flickering scale}
\label{sec:da3lg}
Why does the $4\times$-larger, individually more metric-accurate DA3-g backbone not simply replace DA3-l?
The two models are metrically wrong in different, structurally distinct ways (Fig.~\ref{fig:da3lg}), and
only one of those errors is learnable.

\subsubsection{DA3-l: a stable scale with a large far-field bias}
\label{sec:da3l-bias}
DA3-l is a single monocular network (a DINOv2 ViT-L backbone with a plain DPT depth head) distilled from a
synthetic-data teacher and made metric only by a \emph{deterministic} canonical-focal-length rescale.
Distillation from a smooth synthetic teacher gives it low-variance, temporally stable depth \emph{shape},
but the rescale is a fixed formula rather than a fit; it therefore carries a global scale bias that grows
with range: at ground-truth tracked pixels the median predicted/true depth ratio is $0.56$ on drivetrack overall
and $0.53$ in the far field, against a near-unbiased $0.98$ on adt (Fig.~\subfigref{fig:da3lg}{fig:da3lg-a}). This is precisely the drivetrack
collapse in Table~\ref{tab:abs}---not a tracking failure, but a meter-scale depth bias shared by every
method that unprojects DA3-l depth directly. Because the bias is a smooth, systematic function of range, our
refiner can partially learn it away (drivetrack metric-AJ $0.005\to0.141$), but a single ray cannot fully
recover a $\sim\!2\times$ error at $30$--$100$\,m.

\subsubsection{DA3-g: an unbiased scale with a frame-to-frame flicker}
\label{sec:da3g-flicker}
\begin{figure*}[!tb]
\centering
\begin{subfigure}[b]{0.478\linewidth}
\includegraphics[width=\linewidth]{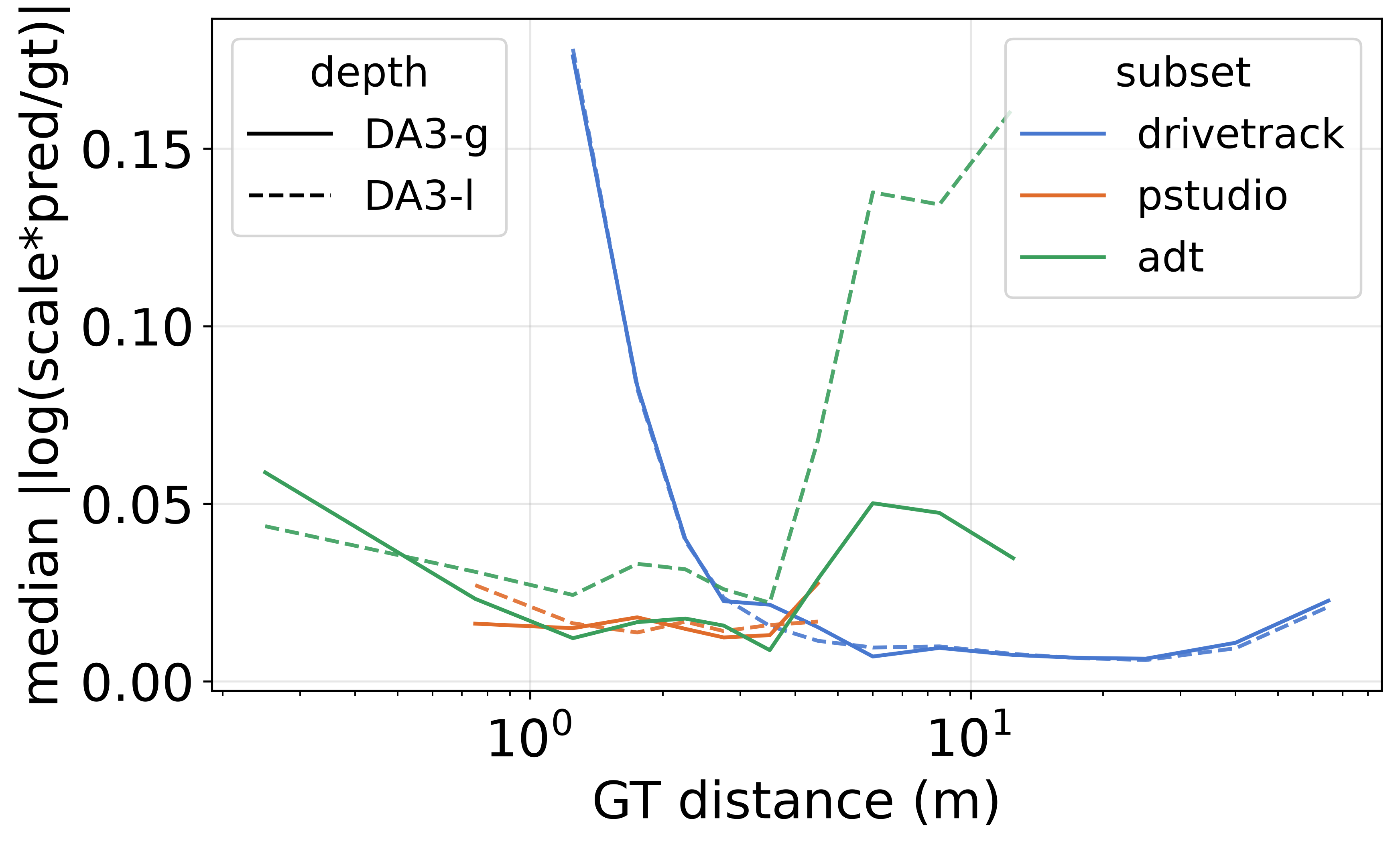}
\subcaption{shape vs GT distance}
\label{fig:da3lg-a}
\end{subfigure}\hfill
\begin{subfigure}[b]{0.478\linewidth}
\includegraphics[width=\linewidth]{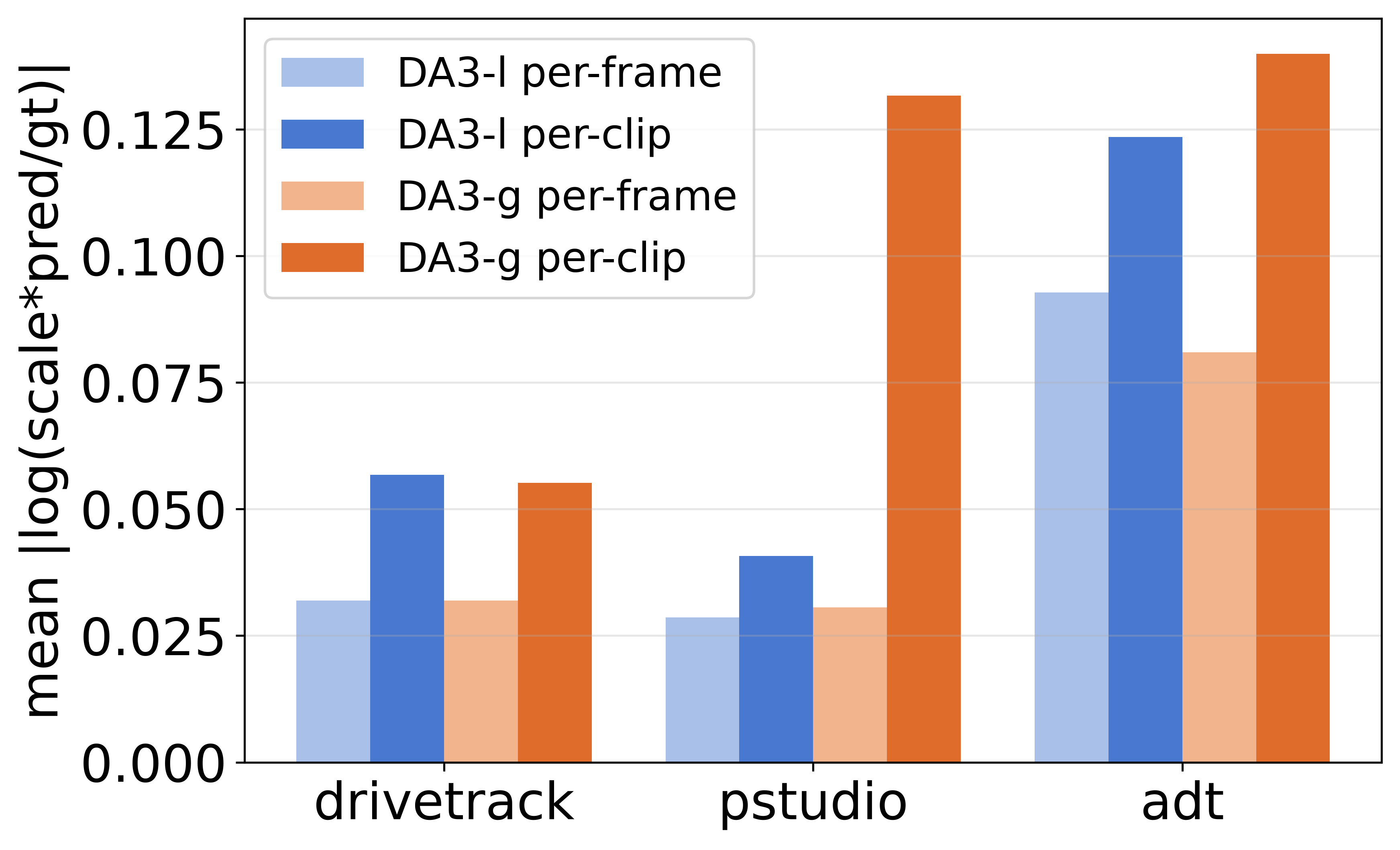}
\subcaption{per-frame vs per-clip scale}
\label{fig:da3lg-b}
\end{subfigure}\\[2pt]
\caption{DA3-l vs.\ DA3-g depth error at ground-truth pixels, TAPVid-3D minival. Here \emph{scale} is
the single positive scalar $s$ by which every depth in a frame (panel a) or a whole clip (panel b) is
multiplied to best match ground truth in a least-squares sense; dividing it out isolates spatial
\emph{shape} error from any error in that one overall multiplier.
DA3-g's \emph{spatial shape} is as good as or better than DA3-l's, but unlike DA3-l, its \emph{scale}
flickers from frame to frame. Throughout, the $y$-axis is a scale-invariant log error,
$|\log(\mathrm{scale}\cdot\mathrm{pred}/\mathrm{gt})|$: $0$ means the prediction exactly matches ground
truth once a scale is removed, and larger values mean more error. \textbf{(a)} Median spatial shape error
vs.\ distance (per-frame scale removed): DA3-g (solid) matches or exceeds DA3-l (dashed) almost everywhere.
\textbf{(b)} Mean error under per-frame vs.\ per-clip scale alignment. Read each backbone across the two
alignments: per-frame $\approx$ per-clip means the scale is temporally stable, and a large gap between
them means the per-frame scale drifts. DA3-l barely changes, whereas DA3-g's pstudio error rises
$4.3\times$ when only one scale per clip is allowed, i.e.\ its per-frame \emph{scale} drifts over time.
The near-field defect is temporal scale instability, not spatial depth noise.}
\label{fig:da3lg}
\end{figure*}

DA3-g nests two branches with two different jobs: an ``any-view'' multi-view-geometry branch (a DINOv2
ViT-g backbone with cross-view attention, trained scale-\emph{invariant} on noisier real multi-view
reconstructions) estimates spatial \emph{shape}, while \emph{metric} scale is estimated only from the DA3-l
branch. The fusion described next is DA3-g's own, internal to the pretrained
checkpoint; we use it as published and never modify it. Critically, it does not fuse the two depths
point-by-point or copy DA3-l's clean depth directly:
it fits one \emph{global scalar} $s$ per frame by least squares so that the any-view branch's shape best
matches DA3-l's depth, $s=\sum_i a_ib_i/\sum_i b_i^2$ ($a{=}$DA3-l depth, $b{=}$any-view depth), then scales
the any-view \emph{shape} by that DA3-l-derived metric scalar and reports $s\times$any-view depth. DA3-l's
own shape is discarded; only that one scalar survives from it. This fixes
the far-field bias (median ratio $1.06$ on drivetrack, including the far field) but introduces a different
defect: the scalar is re-solved \emph{independently every frame} with no coupling to the previous frame, and
because it is a $b^2$-weighted fit it is dominated by the largest (farthest) points visible in that frame.
In a near-field indoor scene the ``far'' pixels are a sparse, occlusion-prone background that changes
frame to frame; the fitted scale therefore wobbles even though its DA3-l target is smooth---the fusion re-solves
the scale each frame rather than inheriting DA3-l's stability.

We isolate this quantitatively at evaluation time only: the model is left untouched, and we instead
remove a single scale from the depth it outputs, in two different ways. \emph{Per-frame} alignment (one
scale per frame) exposes only spatial shape error, while \emph{per-clip} alignment (one scale for the whole
clip) additionally exposes temporal drift; comparing the two separates the one from the other
(Fig.~\subfigref{fig:da3lg}{fig:da3lg-b}). DA3-l's error is nearly identical under
both (pstudio $0.029\to0.041$), confirming a temporally stable scale; DA3-g's near-field error rises
$4.3\times$ under the coarser per-clip alignment (pstudio $0.031\to0.132$), even though its \emph{per-frame
spatial shape} matches or exceeds DA3-l's on every subset (Fig.~\subfigref{fig:da3lg}{fig:da3lg-a}). The defect is therefore a
low-dimensional, one-scalar-per-frame \emph{flicker}, not per-pixel noise.

\subsubsection{Why the 3D refiner cannot repair the flicker}
\label{sec:refiner-no-flicker-fix}
This also explains why our learned refiner, which halves DA3-l's near-field depth error (pstudio median
$0.732\to0.348$\,m), barely moves DA3-g's ($0.777\to0.734$\,m) even though the two raw depths start at
comparable error. A learned corrector can only undo error that is \emph{predictable} from its own inputs.
Our refiner mixes information only along \emph{time}, one track at a time; it never observes a quantity
shared by all points in a frame, and cannot see---let alone correct---a per-frame, all-points-shared
scalar. Oversampling the under-represented near-field training clips and the per-frame scale-stabilisation
stages (Table~\ref{tab:abs-g}) each recover part of this gap by giving the model either more near-field
signal or an explicit within-frame aggregate to estimate the flickering scalar from, but neither fully
closes it. Our full fix pools that shared quantity explicitly, ahead of the refiner rather than inside
it---the two-stage \emph{de-flicker-then-refine} pipeline of \S\ref{sec:scale-refiner}.

\subsubsection{Consequence for backbone choice}
\label{sec:backbone-consequence}
DA3-l's bias is systematic and hence learnable; DA3-g's flicker is a genuine temporal instability that a
per-track model cannot observe, and even the de-flicker stage only partly compensates for it: the
de-flickered DA3-g pipeline (WAFT+DA3-g+de-flicker+vmamba3-2pool, mean absolute metric-AJ $0.224$) still
trails plain DA3-l's $0.248$, and the DA3-l ceiling rises further to $0.255$ once that mixer gains a
second pool and to $0.256$ with the flow-only visibility head. That is why the smaller, biased DA3-l backbone remains our best
\emph{single}-method ceiling even though DA3-g is individually more
metric-accurate at range: the far-field gain DA3-g provides on drivetrack does not offset the near-field
flicker it introduces on pstudio and adt. A depth backbone that combined DA3-l's temporal stability with DA3-g's
far-field metric correctness would outperform both; we leave that fusion to future work.

\subsection{Why external methods underperform under this hardware budget}
\label{sec:reliability}
The low external scores in Table~\ref{tab:median} are genuine consequences of the single-commodity-GPU, pose-free,
monocular budget, which we confirm rather than assume. SpatialTrackerV2 needs a temporal window of $500$
frames ($\sim$$40$\,GB) for its published accuracy; capped at $60$ frames on our GPU it loses most of its
accuracy to windowing drift. TAPIP3D aggregates tracks in a persistent world frame
that requires camera-motion cancellation, which pose-free monocular depth cannot provide; its \emph{own}
official evaluator reproduces our near-zero numbers on the egocentric subsets to three decimals.
TrackCraft3R reports Sim(3)-aligned numbers in its paper; under the standard
per-frame protocol on all $150$ clips it scores far lower, and its video-diffusion backbone runs
$\sim$$200\times$ slower than our baseline. In every case the gap is a property of the method under this
budget, established with the competitor's own tools, not an artefact of our evaluation.

\FloatBarrier
\section{Conclusion}
\label{sec:conclusion}

We set out to track points in \emph{absolute} metric 3D on a commodity single GPU, and we reached that
objective. By freezing optical flow and monocular metric depth and learning only the depth along each point's
fixed pixel ray---conditioned on DINOv3 appearance features and mixed along time by a compact
non-causal state space module of under $1$\,M parameters---the tracker reaches the best \emph{absolute}
metric accuracy on TAPVid-3D minival among all methods evaluated under an identical single-GPU budget,
mean metric-AJ $0.256$ against $0.183$ for the strongest external feed-forward tracker,
while an accompanying reliability analysis, reproduced with competitors' own evaluators, distinguishes
genuine method limits from evaluation artefacts. A depth-backbone analysis further shows that a
$4\times$-larger metric-depth model (DA3-g) trades DA3-l's learnable far-field bias for an unlearnable
frame-to-frame scale flicker; it therefore improves only the far field and does not raise our
single-method ceiling. That analysis motivated the two stages that precede refinement: de-flicker,
which estimates the per-frame scale from the tracked points, and the depth scale refiner, which
estimates it from the depth map directly and reaches our best DA3-g accuracy, mean metric-AJ $0.229$.
The non-causal state space operator the refiner is built from is a means to that result rather than a
claim of its own.

A concrete target for future work is a depth backbone that combines DA3-l's temporal stability with
DA3-g's far-field correctness. A second is a depth model built around vmamba3 from the start and trained
on depth directly, rather than distilled into a frozen DA3: that would let the operator be
evaluated on depth on its own terms, rather than as a swap inside a frozen pipeline.
Replacing DINOv3 with vmamba3 as the tracker's own feature-extraction
backbone, and extending the non-causal cross-attention to full multi-view 3D tracking so that it can observe the
per-frame, all-points-shared quantities a per-track refiner cannot, are natural next steps. More broadly,
non-causal, linear-time attention over 2D and multi-view data is not specific to point
tracking; applying it to other spatiotemporal vision tasks, such as 4D reconstruction, is a promising
direction for future work.


\section*{Data availability}
This study uses the publicly available TAPVid-3D, CIFAR-10, and ETH3D benchmarks; no new data
were generated. The source code and trained models are released at
\url{https://github.com/MasahiroOgawa/visionMamba3} (the vmamba3 operator) and
\url{https://github.com/MasahiroOgawa/vmamba3-3Dpointtracker} (the 3D point-tracking
application); both repositories are made public on publication of this preprint.

\section*{Declaration of generative AI and AI-assisted technologies in the manuscript preparation process}
During the preparation of this work the author(s) used Claude (Anthropic) in order to draft and edit
the text and to write analysis and plotting code. After using this tool, the author(s) reviewed and
edited the content as needed and take(s) full responsibility for the content of the published article.

\appendix
\setcounter{figure}{0}
\setcounter{table}{0}
\section{Standalone behaviour of the non-causal operator}
\label{sec:eval-vm}
\S\ref{sec:vmamba3} derives the non-causal Mamba-3 construction the refiner's mixer is a special
case of, because the tracker depends on it. This appendix records how that operator behaves on its
own, which the tracker does not depend on: the multi-directional scans, the positional encodings and
the rotary analysis below are machinery the pipeline never uses, and the refiners mix over time only.
None of the operator is claimed as novel---Mamba-3~\cite{mamba3}, the multi-directional
scans~\cite{vmamba} and the non-causal collapse~\cite{vssd} are all prior work. What is ours here is:
\begin{enumerate}
  \item \textbf{Their combination}---a single operator in which Mamba-3's three-term
  recurrence, multi-directional scanning and the non-causal collapse are composed and compared as
  interchangeable choices under one implementation and one protocol (\ref{sec:vmamba3-noncausal},
  Table~\ref{tab:cifar}).
  \item \textbf{The extension to cross-attention and multi-view input}: the query side supplies
  $C$ while a separate key/value side supplies $B$ and $V$, Eq.~\eqref{eq:cross}, which is what lets the
  operator mix one track against features drawn from elsewhere rather than against itself
  (\ref{sec:vmamba3-selfcross}).
  \item \textbf{VSSD-2pool}, a second independently-read pool recovering the $\beta$-band capacity
  that the non-causal collapse destroys and that no single per-token vector can hold
  (\ref{sec:vmamba3-noncausal}, construction (iv)).
\end{enumerate}
\noindent together with the measurements reported below. Every configuration is trained once,
under one seed. Where a spread is quoted it is therefore across evaluation views, not across seeds, and
it bounds what one scene can resolve rather than how much a rerun would move.

%
\begin{table*}[!tb]
\centering
\caption{CIFAR-10 from scratch, one fixed training protocol, single seed per cell, at $T{=}65$
         (patch size $4$) and $T{=}1025$ (patch size $1$). Every operator is crossed with
         every positional encoding and their combinations, so that each encoding's own effect can be read separately. \textbf{Enc.} takes one of: ``---'' for no encoding;
         ``2-D RoPE'' for the data-\emph{independent} 2-D rotary embedding on $B,C$
         (\ref{sec:vmamba3-pos}); ``rotary'' for Mamba-3's own data-\emph{dependent}
         complex-SSM rotary; ``1-turn rotary'' for that same rotary with its angular spread
         pinned to one turn instead of growing with $T$; and the two names joined by
         ``$+$'' for the corresponding combination. The 1-turn rows are motivated at
         $T{=}1025$, where the growing form collapses (\ref{sec:rotary-spread}), and are now measured at
         both lengths: pinning costs about five points at $T{=}65$, where the growing rotary is
         still short enough to work, and gains about five at $T{=}1025$.
         Repeat runs of one configuration differed by $0.67$ points; accuracy gaps below
         $\sim$$0.7$ are not meaningful; accuracy is therefore printed to one decimal.
         ``n/a'' marks the CNN's
         $T{=}1025$ cells: a CNN has no notion of tokens and so no sequence length, and is
         reported at $T{=}65$ only. Latency and peak memory are a single forward pass at
         batch $128$ on one RTX\,4080, with the optimiser released, i.e.\ an inference
         footprint.}
\label{tab:cifar}
\small
{\setlength{\tabcolsep}{5pt}
\begin{tabular}{llccccccc}
\toprule
& & & \multicolumn{3}{c}{$T{=}65$} & \multicolumn{3}{c}{$T{=}1025$} \\
\cmidrule(lr){4-6}\cmidrule(lr){7-9}
Operator & Enc. & Params (M) & Acc.\,(\%) & Lat.\,(ms) & Mem (MiB) & Acc.\,(\%) & Lat.\,(ms) & Mem (MiB) \\
\midrule
CNN (ResNet) & --- & 2.78 & 91.8 & 8.9 & 191 & n/a & n/a & n/a \\
\midrule
Softmax attention & --- & 2.69 & 80.9 & 7.7 & 98 & 69.1 & 308.4 & 3592 \\
\midrule
2-dir SSD & --- & 2.71 & 80.4 & 10.0 & 111 & 71.4 & 207.9 & 1306 \\
 & rotary & 2.74 & 81.6 & 11.6 & 135 & 67.5 & 198.2 & 1663 \\
 & 2-D RoPE & 2.71 & 82.8 & 15.8 & 123 & 77.4 & 298.2 & 1498 \\
 & rotary $+$ 2-D RoPE & 2.74 & 81.9 & 15.1 & 147 & 68.0 & 266.8 & 1855 \\
\midrule
4-dir SSD & --- & 2.71 & 81.5 & 16.9 & 129 & 76.1 & 331.7 & 1589 \\
 & rotary & 2.74 & 82.3 & 15.6 & 153 & 82.2 & 342.9 & 1946 \\
 & 2-D RoPE & 2.71 & 83.3 & 19.2 & 129 & 78.5 & 435.3 & 1589 \\
 & rotary $+$ 2-D RoPE & 2.74 & 81.5 & 23.0 & 153 & 63.3 & 420.9 & 1946 \\
\midrule
VSSD-1pool & --- & 2.71 & 75.8 & 9.1 & 97 & 63.9 & 167.4 & 1089 \\
 & rotary & 2.74 & 79.1 & 10.8 & 109 & 19.1 & 222.5 & 1261 \\
 & 1-turn rotary & 2.74 & 70.6 & 9.3 & 126 & 59.0 & 216.1 & 1233 \\
 & 2-D RoPE & 2.71 & 78.4 & 9.8 & 97 & 70.1 & 239.4 & 1089 \\
 & rotary $+$ 2-D RoPE & 2.74 & 81.2 & 13.8 & 109 & 18.9 & 285.4 & 1261 \\
 & 1-turn rotary $+$ 2-D RoPE & 2.74 & 76.3 & 11.7 & 126 & 64.8 & 260.4 & 1233 \\
\midrule
VSSD-2pool & --- & 2.93 & 75.1 & 11.2 & 105 & 66.9 & 213.4 & 1104 \\
 & rotary & 2.97 & 80.0 & 13.7 & 111 & 19.4 & 297.0 & 1263 \\
 & 1-turn rotary & 2.97 & 70.7 & 12.3 & 128 & 59.7 & 315.5 & 1235 \\
 & 2-D RoPE & 2.93 & 79.1 & 16.1 & 99 & 71.4 & 278.5 & 1090 \\
 & rotary $+$ 2-D RoPE & 2.97 & 81.8 & 17.3 & 111 & 18.5 & 408.1 & 1263 \\
 & 1-turn rotary $+$ 2-D RoPE & 2.97 & 75.1 & 15.5 & 128 & 65.4 & 424.0 & 1236 \\
\bottomrule
\end{tabular}}

\\[8pt]
\includegraphics[width=0.92\linewidth]{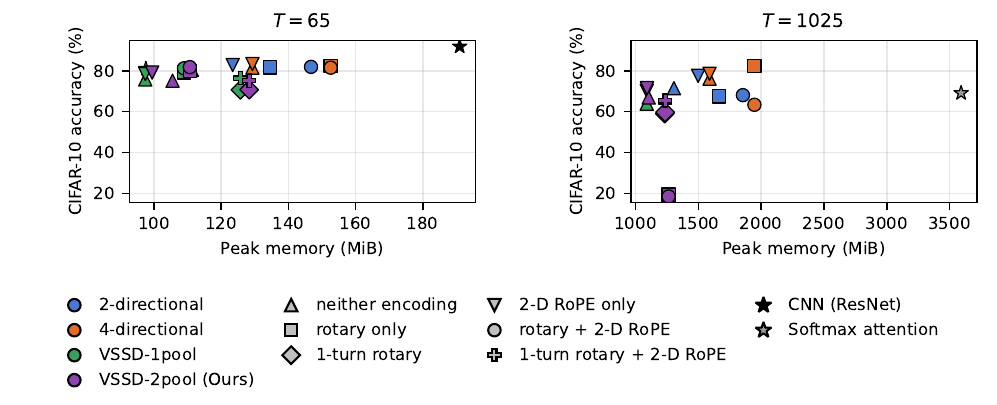}
\captionof{figure}{The same measurements as Table~\ref{tab:cifar}, plotted against peak
memory. Both panels share one accuracy axis so that heights are comparable across sequence
lengths; the memory axes are independent, since peak memory differs by roughly
$20\times$ between them. At $T{=}65$ memory is nearly flat and the plot is effectively
an accuracy ranking; at $T{=}1025$ the state space operators separate from softmax
attention horizontally, which is the advantage the linear-time recurrence provides. The CNN appears
only at $T{=}65$---it does not tokenise and therefore has no token count to place it at on the
right, and its memory does not lie on the same scaling curve.}
\label{fig:cifar-mem-acc}
\end{table*}

\begin{table}[!tb]
  \centering
  \caption{Pinned rotary spread on CIFAR-10 at $T{=}1025$, under Table~\ref{tab:cifar}'s protocol
  (patch size $1$, $30$ epochs, no 2-D RoPE, single seed per cell). $n$ is the total angular spread in
  turns across the sequence, $\theta_j = 2\pi n j/T$. Bold marks the best pinned spread in each column.
  The last three rows are the corresponding cells of Table~\ref{tab:cifar}, repeated for comparison:
  \emph{default rotary} is Mamba-3's own rotary left unmodified, whose angle accumulates to a spread of
  $n\approx160$ at this length (\ref{sec:rotary-spread}); that is why it collapses both operators to near
  chance and is the reason this sweep exists.}
  \label{tab:turns}
  \small
  \begin{tabular}{lcc}
    \toprule
    angular spread & VSSD-1pool & VSSD-2pool \\
    \midrule
    $n=0.1$        & \textbf{66.0} & 65.4 \\
    $n=0.25$       & 65.1 & \textbf{66.6} \\
    $n=0.5$        & 64.7 & 64.8 \\
    $n=1$          & 59.0 & 59.7 \\
    $n=2$          & 58.8 & 59.9 \\
    $n=4$          & 60.6 & 59.4 \\
    $n=10$         & 60.7 & 59.7 \\
    \midrule
    default rotary & 19.1 & 19.4 \\
    rotary off     & 63.9 & 66.9 \\
    2-D RoPE alone & 70.1 & 71.4 \\
    \bottomrule
  \end{tabular}
\end{table}

%
\begin{table}[!tb]
  \centering
  \caption{Metric depth on the held-out ETH3D~\cite{eth3d} \texttt{terrains} scene at matched parameters
  ($\approx22$\,M): DA3-SMALL against the same network with only its self-attention mixer replaced. One
  training run per row, scored on the same twelve views. The three operator rows fall between $0.051$ and
  $0.060$ absolute relative error, a range narrower than the largest within-row standard deviation across
  the twelve evaluated views ($0.017$); they are not separated by this scene and their ordering is not a
  ranking. Bold marks the best
  value in each column. Writing $d$ for predicted and $d^{*}$ for ground-truth depth, and averaging over
  the valid pixels of an image and then over images: abs.\ rel.\ $=\mathrm{mean}\,|d-d^{*}|/d^{*}$, a \emph{relative} error, under which a given metric error counts less at large depth; RMSE
  $=(\mathrm{mean}\,(d-d^{*})^{2})^{1/2}$ in metres, which by squaring weights the worst pixels most;
  $\log_{10}=\mathrm{mean}\,|\log_{10}(d/d^{*})|$, which scores a factor-of-$k$ error equally at every
  depth; and $\delta{<}1.25$, the fraction of pixels with $\max(d/d^{*},d^{*}/d)<1.25$, i.e.\ an inlier
  rate rather than an error. Every row is median-aligned first---$d$ is scaled so that its median matches
  $d^{*}$'s---because monocular depth is recovered only up to a global factor. \emph{The rows did not
  receive the same training.} DA3-SMALL is evaluated zero-shot, with no ETH3D training of any kind,
  whereas each vmamba3 row had a mixer distilled, a bridge trained and the depth head fine-tuned
  on ETH3D (\ref{sec:eval-vm-depth}); their higher $\delta{<}1.25$ follows from that extra training
  and is not evidence of a better backbone.}
  \label{tab:depth}
  \small
  \begin{tabular}{lcccc}
    \toprule
    Mixer & \begin{tabular}{@{}c@{}}abs.\\rel.$\downarrow$\end{tabular}
          & \begin{tabular}{@{}c@{}}RMSE\\(m)$\downarrow$\end{tabular}
          & $\log_{10}\downarrow$
          & \begin{tabular}{@{}c@{}}$\delta{<}$\\$1.25\uparrow$\end{tabular} \\
    \midrule
    softmax (DA3-SMALL)$^{\star}$    & \textbf{0.032} & \textbf{0.061} & \textbf{0.014} & \textbf{1.000} \\
    \midrule
    VSSD-2pool (iv)        & 0.051 & 0.101 & 0.022 & 0.996 \\
    bidirectional (i)               & 0.055 & 0.103 & 0.024 & 0.999 \\
    VSSD-1pool (iii)             & 0.060 & 0.118 & 0.026 & 0.976 \\
    \bottomrule
  \end{tabular}
  \\[2pt]
  {\footnotesize $^{\star}$DA3-SMALL's own attention, evaluated zero-shot with no ETH3D training of
  any kind, its backbone features taken through the same DualDPT path the students use --- not DA3's
  packaged multi-view inference, which additionally selects a reference view and estimates cameras and
  so does not isolate the mixer. Roman numerals index the non-causality constructions of
  \ref{sec:noncausal-constructions}; every row runs with no positional encoding.}
\end{table}

\subsection{Four ways to remove causality}
\label{sec:noncausal-constructions}
\S\ref{sec:vmamba3-noncausal} states which construction the tracker uses and why. This subsection
records the four we considered and compares them under one protocol (Table~\ref{tab:cifar}).
Constructions (i) and (ii) remove causality by scanning the sequence in several directions;
(iii) and (iv) instead collapse the mask to a per-token vector and hold a state whose size does not
grow with the sequence. The tracker mixes along time only and therefore uses a collapse, (iv);
the depth scale refiner is the one exception and uses (iii).

(i) \emph{Bidirectional scans}: run the SSD kernel over forward and reverse token orders and sum them,
giving a \emph{symmetric} mask $\tilde L_{ij}=L^{\rightarrow}_{ij}+L^{\leftarrow}_{ij}$ that depends on
$|i-j|$ rather than $\mathrm{sign}(i-j)$ (Fig.~\ref{fig:vm-attn}(b)), i.e.\ non-causal linear attention at
no extra parameters. Over a raster-flattened image this $|i-j|$ decay still favours whichever axis the
token order happens to serialize as adjacent---row-major puts same-row neighbours $1$ apart but
same-column neighbours $W$ apart---and a single bidirectional pass is therefore dense yet direction-biased.

(ii) \emph{Four-directional scans}, following VMamba's Cross-Scan~\cite{vmamba}: sum a second bidirectional
pair, scanned column-major, on top of the row-major pair of (i)---two symmetric masks added rather than one
(Fig.~\ref{fig:vm-attn}(c)). This removes (i)'s direction bias at no extra parameters, at $2\times$ its
compute: every pixel is now close, in scan order, to both its row and its column neighbours, not just one.
LocalMamba~\cite{localmamba} instead scans inside independent local windows rather than the whole image; we
do not adopt it here, since its own ablations report only a small accuracy gain over plain cross-scanning.

(iii) \emph{VSSD}~\cite{vssd}, which we write \emph{VSSD-1pool}: the query-independent collapse of
Eq.~\eqref{eq:ncssd}, whose $T\times T$ mask reduces to a single $T$-vector $m$ with
$m_j=1/A_j$ (Fig.~\ref{fig:vm-attn}(d)), cutting cost from $O(T^2(N+D))$ to $O(TND)$ with an $O(ND)$
state independent of $T$. The original work names this operator simply VSSD; the
\texttt{1pool} suffix is ours and counts its independently-read pools, distinguishing it from the
VSSD-2pool of (iv), which is not part of~\cite{vssd}. We name these operators by pool count
rather than by the bands of Eq.~\eqref{eq:L3} they were meant to carry, because neither
correspondence is exact: $m_j=1/A_j$ reinterprets the state transition rather than reproducing
$\gamma_j=\lambda_j\Delta_j$, and (iv)'s second vector is freely learned, not constrained to
$\beta$.

(iv) \emph{VSSD-2pool} is derived in \S\ref{sec:vmamba3-noncausal}, being the operator the tracker
is built on.

\subsection{Positional encoding}
\label{sec:vmamba3-pos}
No part of the tracker uses a positional encoding (\S\ref{sec:track-variants}); the three below
exist only for the operator comparison of \ref{sec:eval-vm-classification}, which evaluates every
combination of encoding and operator.
Position information reaches the four operators asymmetrically, which is why the encoding is a separate
choice rather than a fixed part of each. Bidirectional and four-directional scans (i)/(ii) retain an
implicit, decay-based $|i-j|$ bias directly in the mask, with no extra mechanism required.
VSSD-1pool (iii) and VSSD-2pool (iv) collapse the mask to a per-token vector, independent of
the query index; on their own they carry no positional signal at all. We consider three encodings, and
evaluate every combination of them against every operator in Table~\ref{tab:cifar}:

(i) \emph{2-D RoPE}: an ordinary, data-\emph{independent} 2-D rotary position embedding
$\theta_{r,c}[n]=\theta^{\mathrm{row}}[n]\,r+\theta^{\mathrm{col}}[n]\,c$, where $n$ indexes the state
channel pair being rotated. It is a function of the patch's row
and column on the image grid and therefore bounded by the image rather than by the sequence. In the
depth-backbone study (\ref{sec:eval-vm-depth}) this is whichever such embedding the pretrained
Depth-Anything-3 blocks already carry, applied identically whichever operator is swapped in.

(ii) \emph{Rotary}: Mamba-3's own remedy for exactly the collapse above. Promoting $A_t$ from a scalar to
a $2\times2$ rotation is equivalent to a \emph{data-dependent} rotary embedding on $B,C$~\cite{mamba3},
whose angle is a learned per-token increment accumulated along the sequence. It is off by default in our
implementation.

(iii) \emph{Pinned rotary}: the same rotation with its angular spread pinned to $n$ turns across the
sequence, $\theta_j = 2\pi n j/T$, instead of accumulating. This exists because (ii) fails at long sequence
length; \ref{sec:rotary-spread} gives the mechanism and the measurement.

\noindent These encodings only tell a token \emph{where} it is. Which tokens it can see is decided by
(i)--(iv) of \ref{sec:vmamba3-noncausal} alone.

\subsection{Standalone validation: classification}
\label{sec:eval-vm-classification}
We drop the operator into a ViT-Tiny skeleton trained on CIFAR-10~\cite{cifar10} at a matched parameter budget ($\approx2.7$\,M) and compare,
under one fixed training protocol, against softmax attention and a CNN of equal size (Table~\ref{tab:cifar}).

At $T{=}65$ every vmamba3 variant falls within a few points of softmax attention
($80.9\%$), and the best configuration---four-directional with 2-D RoPE, $83.3\%$---exceeds
it by $2.4$, while a well-tuned CNN remains the accuracy ceiling ($91.8\%$). That ceiling is
an expected consequence of CIFAR-10's $50$\,K training images and weak augmentation favouring
a CNN's built-in locality and translation-equivariance over any global-mixing
architecture~\cite{vit}, not evidence that global mixing generalises worse in principle.
Four-directional scanning gives bidirectional's largest plain-operator gain
($80.4\to81.5\%$) at roughly $1.7\times$ its latency, consistent with summing twice as many
scans.

The $2\times2$ of encodings separates each encoding's own effect from the two together.
Every operator improves under either encoding alone, and the two families differ only in the
combination: both VSSD operators are at their best with the rotary and 2-D RoPE on together
($81.2\%$ and $81.8\%$), while both scan operators are at their best with 2-D RoPE alone
and gain nothing from adding the rotary on top (Table~\ref{tab:cifar}). This is what
\ref{sec:vmamba3-pos}'s asymmetry predicts: the collapse carries no positional
signal of its own, and an explicit encoding is therefore worth more to it than to a scan operator,
whose decay band already supplies one.

\paragraph{Long-sequence efficiency} The $T{=}1025$ half of Table~\ref{tab:cifar} is where
the operators' efficiency advantage appears. Softmax attention needs $3592$\,MiB and still scores only
$69.1\%$; every state space configuration runs in $30$--$54\%$ of that memory, and every one
that trains at all exceeds it on accuracy. VSSD-2pool with 2-D RoPE reaches $71.4\%$
at $1090$\,MiB---above VSSD-1pool's $70.1\%$ at the same $1089$\,MiB, which is the lowest
memory in the table---while four-directional with the rotary is the most accurate row at
$82.2\%$ for $1946$\,MiB. Together these are the state space operators' $O(T)$-against-$O(T^2)$
advantage over softmax attention, which is what matters at high resolution and is visible as
horizontal separation in Fig.~\ref{fig:cifar-mem-acc}'s right panel. This is the property the
tracking refiner relies on, since a per-track depth refiner must process clips hundreds of frames
long at a constant per-step memory budget.

One result does not carry over from $T{=}65$, and we report it rather than omit it: the
rotary collapses both VSSD operators to $19.1\%$ and $19.4\%$, barely above the $10\%$
chance floor. The 1-turn rotary repairs the collapse ($59.0\%$ and
$59.7\%$) but still does not reach simply leaving the rotary off ($63.9\%$ and $66.9\%$); at long sequence length we therefore use 2-D RoPE alone.
\ref{sec:rotary-spread} explains the mechanism and the repair.

\subsection{Why a rotary fails at long sequence length, and the repair}
\label{sec:rotary-spread}
The rotary tested here is Mamba-3's own mechanism (\ref{sec:vmamba3-pos}, construction (ii)), run exactly
as its accumulating angle defines it---not a variant we weakened for this comparison. At $T{=}65$ it
helps both VSSD operators. At $T{=}1025$ it reduces them to chance: accuracy falls to $19.1\%$ and
$19.4\%$, barely above the $10\%$ chance floor, while the same encoding makes four-directional the best
row in the table (Table~\ref{tab:cifar}). The failure is therefore specific to the collapse, and its
cause is the pooled state the collapse creates, not a defect we introduced into the rotary itself.

A rotary encodes position by rotating $B$ and $C$ by an angle that accumulates along the sequence. A scan
operator forms the score $C_i^{\top}B_j$ one pair at a time; the two absolute angles therefore cancel and
only their difference $\theta_i-\theta_j$ matters; that difference does not grow with $T$. VSSD instead adds
the keys into a single pooled state, $\sum_j m_j\,\mathcal{R}(\theta_j)B_j$, before any query reads it.
Vectors rotated by very different angles point in unrelated directions, and adding them cancels them out.
Because $\theta$ is a plain cumulative sum, the spread of angles grows with the sequence: about ten full
turns at $T{=}65$, but about $160$ at $T{=}1025$. At $160$ turns the pooled state is already close to
random before training begins, which is why accuracy remains at chance.

The repair is to fix the total spread rather than let it accumulate. Setting
$\theta_j = 2\pi n j/T$ makes the sequence span exactly $n$ turns whatever $T$ is, keeping neighbouring tokens
distinguishable while the sum no longer cancels. At $n{=}1$ both operators recover from chance
to about $59\%$.

One turn is not, however, the best choice. Sweeping $n$ on both VSSD operators under
Table~\ref{tab:cifar}'s protocol gives Table~\ref{tab:turns}.

\noindent The sweep establishes a threshold rather than a trend. For both operators every spread of one
turn or more lies between $58.8\%$ and $60.7\%$, while every spread of half a turn or less lies between
$64.7\%$ and $66.6\%$; the threshold replicates on both, which is what the pooled-state argument predicts.
Two things the sweep does not establish are worth stating. Above one turn the ordering does not
replicate---VSSD-1pool is highest at ten turns and VSSD-2pool at two, across a range under
$2$ points wide---and the residual differences there are within run-to-run variation. Below half
a turn the two operators disagree about where the best value lies: VSSD-1pool continues to improve down
to a tenth of a turn ($66.0\%$) while VSSD-2pool peaks at a quarter turn ($66.6\%$) and then
declines. We therefore report only that the spread should be well below one turn, and do not recommend a
particular $n$.

Rescaling a rotary's angle is not our idea: position interpolation~\cite{posinterp} and
YaRN~\cite{yarn} already do it. The difference is why. They shrink the angle so that a model can run on
sequences longer than the ones it was trained on. We shrink it at a fixed length, to stop the pooled sum
from cancelling.

Repairing the collapse does not make the rotary preferable. At its best pinned spread VSSD-1pool
reaches $66.0\%$ against $63.9\%$ with the rotary switched off, and VSSD-2pool $66.6\%$ against
$66.9\%$: a small gain in one case and a tie in the other. Both stay well below 2-D RoPE alone
($70.1\%$ and $71.4\%$), whose angle depends on the patch's row and column and is therefore bounded by the image
rather than growing with the sequence. The tracking pipeline therefore uses no rotary of either kind.

\subsection{Standalone validation: metric depth estimation}
\label{sec:eval-vm-depth}
The second standalone task tests whether the operator can replace softmax attention inside a
\emph{pretrained} depth network. We take DA3-SMALL, replace only its self-attention mixer layers with a
vmamba3 operator at matched parameter count ($\approx22$\,M), and retain DA3's own DualDPT depth
head.

The substitution requires one architectural addition, because DA3's depth head expects
$2\,d_{\text{embed}}$ features from its concatenated local and global streams whereas our backbone emits a
single $d_{\text{embed}}$ stream. Instead of duplicating that stream, we interpose a learnable
\emph{bridge} $\mathbb{R}^{d_{\text{embed}}}\!\rightarrow\!\mathbb{R}^{2d_{\text{embed}}}$, initialised to
the duplication and trained during fine-tuning, which is what allows the pretrained head to be reused with
a different mixer.

Training proceeds in three stages, all on a $374$-image, $10$-scene subset of ETH3D~\cite{eth3d}, with the
\texttt{terrains} scene held out for evaluation. The mixer is first distilled for $20\,000$ steps against
the intermediate features of the frozen DA3-SMALL teacher at four backbone layers, under a combined L2 and
cosine objective. Depth is then supervised against ground truth for $500$ steps with the head frozen, and
for a further $1000$ steps with the head unfrozen at a reduced learning rate and with augmentation
enabled. The two depth stages are kept separate because the second begins from an already-adapted bridge.
All three stages use AdamW~\cite{adamw} with weight decay $0.05$, under a cosine schedule floored
at a tenth of the peak learning rate rather than annealed to zero.
Both operators are trained without positional encoding; this configuration therefore differs from the 2-D
RoPE variant that \ref{sec:eval-vm-classification} identifies as preferable at long sequence length.

Table~\ref{tab:depth} reports all three non-causal operators against the DA3-SMALL baseline, every row
trained and scored with the released code and all four scored by one evaluator. The three operators fall between
$0.051$ and $0.060$ absolute relative error. That range is narrower than the largest within-row standard
deviation across the twelve evaluated views ($0.017$, on VSSD-1pool); this scene does not separate them
and their ordering is not a ranking: VSSD-2pool is best on absolute relative error, RMSE and
$\log_{10}$ and bidirectional on the inlier rate, by margins this experiment cannot establish as
real. All three are roughly $1.6$ to $1.9$ times DA3-SMALL's error. That is
the substantive outcome: substituting the operator into a pretrained depth pipeline costs accuracy rather
than improving it, and the adaptation described above recovers part but not all of what it costs.

The comparison is not a controlled one, and it favours each side in a different respect. DA3-SMALL is
evaluated zero-shot, with no ETH3D adaptation of any kind, whereas every vmamba3 row has had a
mixer distilled, a bridge trained and a head fine-tuned on the evaluation dataset; their higher
$\delta{<}1.25$ inlier rates are consequently not evidence of a superior backbone. Conversely, DA3-SMALL
benefits from the whole of DA3's pretraining corpus, whereas our mixers are initialised randomly in their
attention layers and trained on $374$ images. The experiment therefore supports only a narrow claim: a
non-causal state space mixer \emph{can} be substituted for softmax self-attention within an
already-pretrained depth pipeline at $1.6$ to $1.9$ times its error (Table~\ref{tab:depth}). A stronger conclusion would require
DA3-SMALL to be given the same adaptation budget, and our mixers to be pretrained at comparable scale;
neither experiment has been performed.

One limit is worth stating rather than leaving to be inferred: this is a single seed per row over twelve
evaluated views of one held-out scene. The per-view spread above bounds what it can resolve, and every
difference between the three operators falls inside that bound. Separating them would need more scenes,
more seeds, or both.

\subsection{What the two standalone studies establish}
\label{sec:eval-vm-effectiveness}
CIFAR-10 and ETH3D are not the same kind of test and should not be read the same way, and only CIFAR-10
is a controlled one: it is unconstrained, with no bias toward either operator, and every cell is trained
under one fixed protocol. ETH3D is a drop-in-replacement test whose rows differ in adaptation as well as
in operator (\ref{sec:eval-vm-depth}), and its three operators fall within the per-view spread of one
another; it therefore establishes that the substitution is viable without ranking them. CIFAR-10 covers all four
operators but at $32\times32$ resolution on a classification objective.
The claim they support therefore rests on CIFAR-10: vmamba3 provides efficiency
and long-context scalability at accuracy parity. That is the property the tracker of \S\ref{sec:method}
draws on, and the only one it needs.

\FloatBarrier

\bibliographystyle{elsarticle-num}
\bibliography{refs}

\end{document}